%% file: main.tex
\documentclass{article}

\usepackage[final]{corl_2020} 

\usepackage{amsfonts}

\input{packages}
\input{macros}

\input{math_commands}

\title{A Long Horizon Planning Framework for Manipulating Rigid Pointcloud Objects}

\author{
    Anthony Simeonov$^1$,
    Yilun Du$^1$,
    Beomjoon Kim$^{1,2}$,
    Francois R. Hogan$^{1,3}$,\\
    \textbf{Joshua Tenenbaum}$^1$,
    \textbf{Pulkit Agrawal}$^1$, 
    \textbf{Alberto Rodriguez}$^1$ \\
    $^1$Massachusetts Institute of Technology --- \texttt{\{asimeono,yilundu,jbt,pulkitag,albertor\}@mit.edu} \\
    $^2$KAIST Graduate School of AI --- \texttt{beomjoon.kim@kaist.ac.kr}\\
    $^3$Samsung AI Center Montreal --- \texttt{f.hogan@samsung.com}
}

\begin{document}
\maketitle

\input{text/abstract}

\keywords{Manipulation, Learning, Planning} 


\input{text/introduction_as}

\input{text/problem_setup_rev1} 
\input{text/method_rev1}


\input{text/experiment}
\input{text/related_work}
\input{text/conclusion}


\clearpage

\input{text/acknowledgement}


{\small \bibliography{include/example.bib}}

\input{text/appendix}

\end{document}

%% file: packages.tex
\usepackage{algpseudocode}
\usepackage{algorithmicx}
\usepackage[ruled]{algorithm}

\usepackage{color,xcolor}
\usepackage{epsfig}
\usepackage{graphicx}

\usepackage{adjustbox}
\usepackage{array}
\usepackage{booktabs}
\usepackage{colortbl}
\usepackage{float,wrapfig}
\usepackage{hhline}
\usepackage{multirow}
\usepackage{subcaption} 

\usepackage{amsmath,amsfonts,amsthm,amssymb}
\usepackage{bm}
\usepackage{nicefrac}
\usepackage{microtype}

\usepackage{changepage}
\usepackage{extramarks}
\usepackage{fancyhdr}
\usepackage{lastpage}
\usepackage{setspace}
\usepackage{soul}
\usepackage{xspace}

\usepackage{url}

\usepackage{enumerate}
\usepackage{todonotes} 

\usepackage{titlesec}
\usepackage{makecell}

\usepackage{amsmath,amsthm,amssymb}
\usepackage{mathtools}
\usepackage{stmaryrd}
\usepackage{enumitem}

%% file: macros.tex
\newcolumntype{L}[1]{>{\raggedright\let\newline\\\arraybackslash\hspace{0pt}}m{#1}}
\newcolumntype{C}[1]{>{\centering\let\newline\\\arraybackslash\hspace{0pt}}m{#1}}
\newcolumntype{R}[1]{>{\raggedleft\let\newline\\\arraybackslash\hspace{0pt}}m{#1}}

\newcommand{\sect}[1]{Section~\ref{#1}}

\newcommand{\ignore}[1]{}

\makeatletter
\DeclareRobustCommand\onedot{\futurelet\@let@token\@onedot}
\def\@onedot{\ifx\@let@token.\else.\null\fi\xspace}

\g@addto@macro\normalsize{%
  \setlength\abovedisplayskip{0pt}
  \setlength\belowdisplayskip{0pt}
  \setlength\abovedisplayshortskip{0pt}
  \setlength\belowdisplayshortskip{0pt}
}
\makeatother

\definecolor{MyDarkBlue}{rgb}{0,0.08,1}
\definecolor{MyDarkGreen}{rgb}{0.02,0.6,0.02}
\definecolor{MyDarkRed}{rgb}{0.8,0.02,0.02}
\definecolor{MyDarkOrange}{rgb}{0.40,0.2,0.02}
\definecolor{MyPurple}{RGB}{111,0,255}
\definecolor{MyRed}{rgb}{1.0,0.0,0.0}
\definecolor{MyGold}{rgb}{0.75,0.6,0.12}
\definecolor{MyDarkgray}{rgb}{0.66, 0.66, 0.66}

\DeclareRobustCommand{\asnote}[1]{\ifthenelse{\boolean{draft-mode}}{\textcolor{blue}{\textbf{AS: #1}}}{}}
\DeclareRobustCommand{\arnote}[1]{\ifthenelse{\boolean{draft-mode}}{\textcolor{green}{\textbf{AR: #1}}}{}}

\newcommand{\myparagraph}[1]{\vspace{-6pt}\paragraph{#1}}
\newcommand{\myitem}{\vspace{-2pt}\item}

%% file: math_commands.tex
\usepackage{amsmath,amsfonts,bm}

\def\eqref#1{equation~\ref{#1}}

\def\1{\bm{1}}

\DeclareMathAlphabet{\mathsfit}{\encodingdefault}{\sfdefault}{m}{sl}
\SetMathAlphabet{\mathsfit}{bold}{\encodingdefault}{\sfdefault}{bx}{n}

\DeclarePairedDelimiterX{\infdivx}[2]{(}{)}{%
  #1\;\delimsize|\delimsize|\;#2%
}

%% file: text/abstract.tex
\begin{abstract}
We present a framework for solving long-horizon planning problems involving manipulation of rigid objects that operates directly from a point-cloud observation, i.e. without prior object models. Our method plans in the space of object subgoals and frees the planner from reasoning about robot-object interaction dynamics by relying on a set of generalizable manipulation primitives. We show that for rigid bodies, this abstraction can be realized using low-level manipulation skills that maintain sticking contact with the object and represent subgoals as 3D transformations. To enable generalization to unseen objects and improve planning performance, we propose a novel way of representing subgoals for rigid-body manipulation and a graph-attention based neural network architecture for processing point-cloud inputs. We experimentally validate these choices using simulated and real-world experiments on the YuMi robot. Results demonstrate that our method can successfully manipulate new objects into target configurations requiring long-term planning. Overall, our framework realizes the best of the worlds of task-and-motion planning (TAMP) and learning-based approaches. Project website: \url{https://anthonysimeonov.github.io/rpo-planning-framework/}.
\end{abstract}

%% file: text/introduction_as.tex
\section{Introduction}
\label{sec: introduction}
Consider the bi-manual robot in Figure~\ref{fig: intro} tasked with moving a block (red) into a target pose on the far side of the table (green). The robot has a library of manipulation skills: \textit{pull}, \textit{push} and \textit{grasp-reorient} (flips the block so it rests on a different face). To complete the task, the robot must first pull the object near the center of the table, where it can be grasped with both palms to reorient it. After reorientation, the robot cannot directly place the block at the target position due to its limited reach. Instead, the block must first be placed within reach of both manipulators and then pulled to the target by the left hand. We aim to develop a robotic system that addresses the perceptual and planning challenges of such long-term tasks requiring coordinated and sequential execution of multiple skills on a variety of objects of different shapes and sizes. Two key observations guide the approach we propose:

\begin{itemize}[leftmargin=*]
    \myitem The challenge of \textbf{planning for long-term tasks} arises from a large search space. Instead of reasoning in the space of low-level robot actions, the search space can be substantially reduced by planning in the more \emph{abstract} space of key object configurations (or, \emph{subgoals}) -- an idea well studied in Task-and-Motion planning (TAMP)~\cite{LIS134,CambonIJRR2009,GarrettIJRR2017,ToussaintIJCAI2014}. However, the challenge is that some subgoals may be unachievable either due to kinematic/dynamic constraints of the robot or because the low-level skill-set may be incapable of achieving a particular subgoal. Because all subgoals are not feasible, it is not effective to decouple high-level planning of \textit{what} the subgoals are and the low-level control of \textit{how} the robot reaches these subgoals.     
    Decoupling necessitates many explicit feasibility checks and leads to severe computational inefficiencies from searching through many infeasible plans. We mitigate this inefficiency by simultaneously biasing search toward subgoals that are reachable and predicting both subgoals and how to achieve them. 
    \myitem The challenge of \textbf{perceiving unknown objects of diverse shapes and sizes} arises due to the unavailability of 3D models of novel objects and errors in reconstructing an object's full geometry/state from noisy sensor observations. 
    We adopt the alternative approach of building perception models that directly predict skill parameters from raw sensory observations. This idea has been successfully used by affordance-prediction approaches to manipulation~\cite{mahler2017dex, hermans2011affordance,zeng2018arc,manuelli2019kpam,kloss2019accurate}.
    In our case, for \emph{every skill}, we learn a separate \emph{skill parameter sampler} that predicts a distribution over \emph{subgoals} and \emph{contact locations} that encode how to achieve the subgoals directly from an object point-cloud.
\end{itemize}
\input{figText/intro}

Conventional TAMP methods construct long-term plans, but they often rely on hand-designed heuristics for sampling subgoals. These heuristics are typically based on \textit{a priori} information about the object's shape, limiting their utility when attempting to manipulate novel objects. Alternatively, prior works performing object manipulation directly from sensory observations have shown generalization to new objects~\cite{agrawal2016learning, finn2017deep, ebert2018visual}, but face difficulty in long-horizon tasks because they operate in the space of low-level actions. We can overcome this short-sightedness while maintaining a similar ability to generalize by predicting subgoals from sensory observations. However, for efficient planning it is crucial to constrain the predicted subgoals to be feasible and also find actions for achieving them. 

Herein lies our central problem: to perform \emph{efficient} long-horizon manipulation planning with a variety of unknown objects directly from \emph{sensory observations}. In this paper, we present a framework that addresses this problem for quasi-static rigid body manipulation. Planning in the space of subgoals would be efficient if the low-level controller can easily achieve them. For rigid bodies, $SE(3)$ transformations describe the space of all possible subgoals. Now, if we further assume that the relative transformation between the object and the robot remains fixed during manipulation (i.e., the commonly made sticking-contact assumption~\citep{chavan-dafle2020,hogan2020tactile}), the desired transformation of the object is the same as the transformation that the manipulator must execute after it makes contact with the object. This means that the problem of predicting an action sequence to achieve the subgoal can be reduced to simply predicting the initial contact configuration with the object. 

Given an object subgoal and the contact information, inverse kinematics can be used to construct the low-level controller. The challenge comes from maintaining sticking contact during manipulation, a problem that was recently solved by~\citet{hogan2020tactile} to produce a library of manipulation skills for pulling, pushing and re-orienting cuboidal objects. However, their approach requires the object's 3D model. To generalize these skills to work directly from point-clouds, we use self-supervised learning wherein the robot first uses privileged information to manipulate cuboidal objects using these skills. This provides training data that maps a point-cloud observation of the object to a \emph{feasible} set of transformations that can be applied (i.e., subgoals expressed as rigid object transformations) and the contact configuration (i.e., cartesian poses of the end effector) required to achieve the subgoal. Using this data, we learn a deep ``sampler" that maps an object point-cloud to compatible distributions of both feasible subgoals and contact locations, which enables long-term planning. 



Overall, our framework generalizes traditional TAMP approaches to work with perceptual representations and previously unseen objects while still enabling long-term planning, which remains challenging for end-to-end learning-based systems.
Our technical contributions are: (i) a planning and perception framework for sequential manipulation of rigid bodies using point-cloud observations;
(ii) specific architectural choices that significantly improve planning performance and efficiency. These choices include: (a) an object/environment geometry-grounded representation of reorientation subgoals for neural networks to predict; (b) a novel graph-attention based model to encode point-clouds into a latent feature space; (c) and joint modeling of what subgoals to achieve and how to achieve them, instead of modeling them independently. 

We validate our approach using simulated and real-world experiments on a bi-manual ABB YuMi robot that manipulates objects of previously unseen geometries into target configurations by sequencing multiple skills. We show our method outperforms meticulously designed baselines that utilize privileged knowledge about object geometry and task in terms of planning efficiency. We present detailed ablation studies to quantify the performance benefits of our specific system design choices. 

%% file: figText/intro.tex
\begin{figure}[t]
    \centering
    \includegraphics[width=\linewidth]{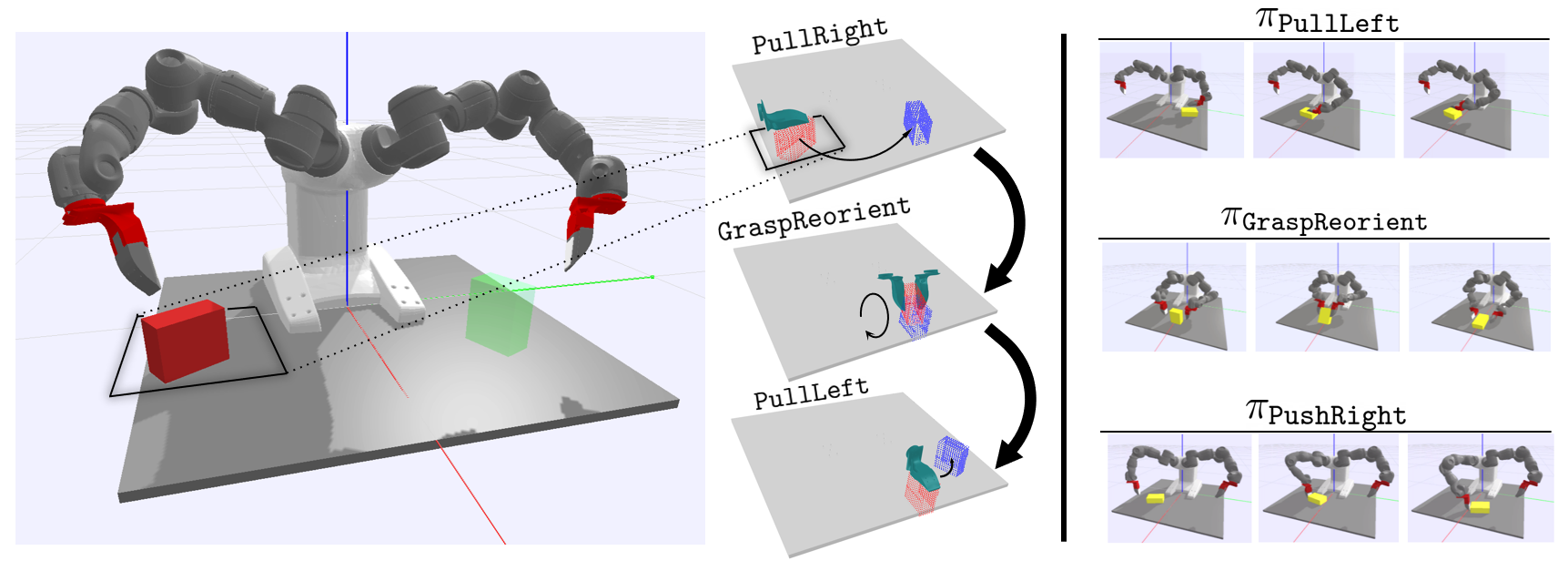}
    \caption{{\bf Left}: Our framework uses learned samplers and a family of primitive manipulation skills to imagine and execute multi-step plans that manipulate objects between stable poses, using only a segmented object point-cloud. First (top right), our learned model for the \textit{pull} skill samples a palm pose (dark green) and a subgoal (blue), when provided with an input point-cloud (red). The \textit{pull} subgoal becomes the input at step two for \textit{grasp-reorient}. Sampling repeats iteratively toward the goal (green).  {\bf Right}: Examples executions of the manipulation skills from \cite{hogan2020tactile} we use in our setup.} 
    \label{fig: intro}
    \vspace{-10pt}
\end{figure}

%% file: text/problem_setup_rev1.tex
\section{Problem Setup and Multi-step Planning}
\label{sec: problem-setup}
\label{sec: multi-step-planning}

\myparagraph{Assumptions} 

Our work makes three key assumptions: 
1) Objects are rigid and are represented as a segment of a point-cloud, which we refer to as \emph{Rigid Pointcloud Objects}; 
2) Subgoals are represented as sequences of $SE(3)$ transformations of the object point-cloud; and 
3) The object maintains a fixed configuration with respect to the robot during interaction due to an assumed sticking contact.
We also assume quasi-static interactions, segmentation of the entire point-cloud to obtain the object point-cloud, and a plan skeleton~\cite{lozano2014constraint} that specifies the sequence of skill \textit{types} for solving the task. Note that assuming the plan skeleton is not restrictive, as a variety of techniques have been proposed to obtain it~\cite{garret17skeleton, beomjoon2019relationalGTAMP}, and recent work demonstrated it can be predicted from sensory observations~\cite{driess2020deep}. 

\myparagraph{Task Setup}
Given a segmented point-cloud observation of a rigid object, $X \in \mathbb{R}^{N \times 3}$, the task is to transform the  object by an $SE(3)$ transformation denoted by $T^o_{des}$. It is assumed that the target configuration can be reached in $T$ steps. The planner must determine a sequence of object transformations, $T^{o}_{1:T} \in SE(3)$ and a sequence of poses of the left and right manipulator when they make contact with the object denoted as $T^{p_{c}}_{1:T} = \{T^{p_{c}}_{L_{1:T}, R_{1:T}}\} \in SE(3)$, where $T^{p_{c}}_{L_{i}, R_{i}}$ denotes the pose of the left and right manipulator respectively at the $i^{th}$ step of the plan. If the plan is successful, then $\prod_{i=1}^{T}T^o_{i}= T^o_{des}$. Under the sticking contact assumption, planning in the space of $(T^{p_c}, T^o)$ is sufficient for rigid body manipulation. 
Sticking contact ensures that desired object transformation $T^{o}_i$ can be achieved by moving the manipulators by the same transformation  ($T^{o}_i$) after they make contact with the object. We denote $T^{p_c}$ as a contact pose and $T^o$ as a subgoal.

\myparagraph{Manipulation Skills}
From the work of~\cite{hogan2020tactile}, we re-use a set of six (K=6) manipulation skills $\{\pi^1,...,\pi^K\}$ that transform the object while maintaining sticking contact. The skills are pulling (R/L), pushing (R/L), pick-and-place, and re-orientation~\cite{hogan2020tactile} (Figure \ref{fig: intro} Right). 
Given $(T^{p_c}, T^o)$ determined by the planner, each skill operates as a low-level controller, 
\begin{equation}
    \label{eq:skills}
    \mathbf{q}_{R, L} = \pi(T^{p_c}, T^o)    
\end{equation}
that outputs a sequence of joint configurations of the right $(\mathbf{q}_{R})$ and left $(\mathbf{q}_{L})$ manipulators required to transform the object by $T^o$. The combined joint sequence is denoted by $\mathbf{q}_{R,L}= \{q_0,...,q_F\}$. The sequence length ($F$) is variable and depends on the skill-type and the inputs to $\pi$. The skill is feasible if $\mathbf{q}_{R,L}$ is collision-free, respects joint limits, and avoids singularities.

\myparagraph{Planning} We assume that the robot is given a plan skeleton $PS = \{S_t\}_{t=1}^{T}$, where $S_t \in \pi^{1:K}$ denotes the skill-type. Given this skeleton, the planning problem is to determine for every skill the: (a) object subgoal transformation ($T^o_{t}$) and (b) contact poses ($T^{p_c}_t$). We use a sampling-based planner to search for sequences of object transformations and contact points $\{T^{o}_{1:T}, T^{p_c}_{1:T}\}$. Each skill $(S_t)$ in the plan skeleton ($PS$) can be thought of as a node and the goal of planning is to connect these nodes. Application of skill $S_t$ at node $t$ transforms the input point-cloud $X_{t}$ into the next point-cloud $X_{t+1} = T^{o}_{t}X_t$. To efficiently sample plausible plans, it is necessary to only sample $T^{o}_t$ that can be achieved by the $S_t$ from the current object configuration $X_t$. For this, we learn a sampler for the subgoal transformations and contacts, $(T^o_t, T^{p_c}_t)$, using a neural network (NN), denoted $p_{\pi_t}(\cdot|X_t; \theta)$, where $\theta$ are weights of the NN. 
Because the sampler is imperfect, we may still encounter infeasible $( T^o_t, T^{p_c}_t)$, which are discarded. 
For the final step in $PS$, $T^o_T$ can be analytically solved as $T^o_T = T^o_{des}\big(\prod_{t=0}^{T-1}T^o_{t}\big)^{-1}$. To realize $PS$, the sampling procedure is run either until we find a feasible sequence $(T^{p_c}_{1:T}, T^o_{1:T})$ or the planner times out. See Appendix~\ref{sec: app-multistep-planning-algo} for details on the full algorithm.

%% file: text/method_rev1.tex
\section{Learning the Skill Sampler}
\vspace{-5pt}
\input{figText/network_arch}
\label{sec: methods}
We represent skill samplers $\{p_{\pi^i}(\cdot | X ; \theta)\}_{i=1}^{K}$ using a conditional variational autoencoder (CVAE;~\cite{sohn2015cvae}) that predicts a distribution over skill parameter $(T^{p_c}, T^o)$ from a point-cloud observation (see Appendix~\ref{sec: app-cvae-sampler} for details). The CVAE is trained using a dataset $D_{\pi^i}$ of \emph{successful} single-step skill executions, $D_{\pi^i} = \{(X^{(i)}, T^{o~(i)}, T^{p_c~(i)}) \}_{i=1}^{n}$, which ensures that the predicted samples are very likely to be feasible. This makes high-level planning by sampling from $p_{\pi^i}$ very efficient, while maintaining the ability to generalize that comes from operating directly from sensory observations. We now describe the specific modeling choices that were critical for performance.

\myparagraph{Learning Pointcloud Features for Predicting Contact Poses}
Predicting the end-effector pose at which the robot contacts the object $T^{p_c}$ and the subgoal $T^{o}$ requires a feature representation that accurately models the object geometry. Prior work used models such as  PointNet++\cite{qi2017pointnet++} for successfully predicting end-effector poses \cite{mousavian2019graspnet, murali2019collisionnet, qin2019keto, fang2019cavin}. However, PointNet++ style architectures only coarsely model the object geometry, which results in poor performance when accurate geometry-aware predictions must be made. E.g., in our setup, accurate alignment of left and right grippers is necessary to manipulate the object. To capture rich geometric shape information, we hypothesized that it was necessary to explicitly model pairwise relationships between points in the point-cloud. For this, we made use of Graph Attention (GAT)~\cite{velivckovic2017graph} in the decoder architecture, shown in Figure~\ref{fig:network} (Middle), that outputs contact end-effector poses and the object subgoal. Each point is treated as a node in a fully-connected graph and represented by a concatenated feature vector: $(p_j \oplus z ); j \in [1,N]$, where $z$ is a sample from the learned latent distribution and $p_j$ is the 3D coordinate of the $j^{th}$ point. After processing the inputs with multiple rounds of self-attention, two separate branches of fully-connected layers predict $T^{p_c}$ and $T^{o}$, respectively. While average pooling of node features sufficed to predict $T^{o}$, separately predicting $T^{p_c}$ for each node improved learning performance. Accurate predictions from single node features is only possible if the features capture global geometric information. Forcing each node to individually predict $T^{p_c}$ therefore leads to learning of superior geometry-aware features. At test time, the average of the $T^{p_c}$ predictions is used.

\label{sec:mask-representation}
\myparagraph{Obtaining Accurate Subgoal Prediction}
\input{figText/pipeline}
In our setup, some skills such as \textit{pull} move objects in $SE(2)$ and other skills such as \textit{grasp-reorient} move the object in $SE(3)$. We found that while our model made accurate predictions of $SE(2)$ transformations, directly predicting $SE(3)$ transformations represented as translations and rotations (expressed using quaternions) led to inaccurate predictions. The inaccuracies in $T^o$ predictions can easily accumulate over a long planning horizon and result in either poor planning performance or termination of planning due to predictions of physically unrealizable configurations, such as the object floating in the air, penetrating the table, etc. (Figure~\ref{fig:infeasible-subgoals-predicted-transform} shows such examples). We resolve this problem by leveraging the insight that predicting the reorientation subgoal (i.e., $SE(3)$ transformation) can be recast as the problem of selecting \textit{which points on the object should end in contact with the support surface}. To implement this, we compute the rotation component of $T^{o}$ by first predicting for every point in the object point-cloud, the probability of the point resting on the table after $T^{o}$ is applied. The per-point probabilities are then thresholded to obtain a binary segmentation mask $\{X_{mask}:  \{m_j \in \{0, 1\} \forall j \in [1,N]\}\}$. Next, we use Iterative Closest Point (ICP)~\cite{paul1992beslICP} to perform registration of the table point-cloud $X_{table}$ and $X_{mask}$ to solve for the orientation (see Figure~\ref{fig:full-pipelines}). This inferred orientation is combined with the predicted translation to obtain $T^o$. This pipeline enables a more accurate subgoal prediction because the transformation that is obtained is grounded to represent an explicit spatial relationship between points on the object and points on the table.  

\myparagraph{Modeling the joint distribution of Subgoals and Contact Points}
Independent modeling of subgoals and contact poses $(T^o, T^{p_c})$ can lead to incompatible predictions. E.g., a sampled $T^{p_c}$ can result in a robust sticking contact, but the motion required to achieve an independently sampled subgoal $T^{o}$ can result in the robot colliding with the table. We mitigate this issue by modeling the joint distribution of $(T^o, T^{p_c})$ by training the CVAE encoder to learn a single shared latent distribution. The decoder uses this latent representation to predict $T^o, T^{p_c}$ and $X_{mask}$ using a shared set of self-attention layers and separate output heads. By training the model to predict the skill parameters jointly using weights and latent samples which are shared, we can more reliably obtain subgoals and contact poses that are compatible with each other.

\myparagraph{Training Data}
The training data $D$ is generated by manipulating cuboids of different sizes using one primitive skill at a time. For data collection, we sample cuboids so that they are in a stable configuration on the table. We then randomly sample subgoals $(T^{o~(i)})$ and contact poses $(T^{p_c~(i)})$ on the object mesh. Assuming knowledge of the 3D models of the cuboids, we execute a single primitive skill~\citep{hogan2020tactile}. If the skill successfully moves the object to the subgoal, we add the tuple of (object point-cloud, contact pose, subgoal, and $X_{mask}$) to the dataset. We used manipulation data from 50-200 different cuboids for each skill, and our overall dataset contained 24K samples. We trained separate CVAEs with the same architecture for each skill. See Appendix~\ref{sec: app-model-architecture-details} for more information on data generation, training, and network architecture.

%% file: figText/network_arch.tex
\begin{figure}
    \centering
    \includegraphics[width=\linewidth]{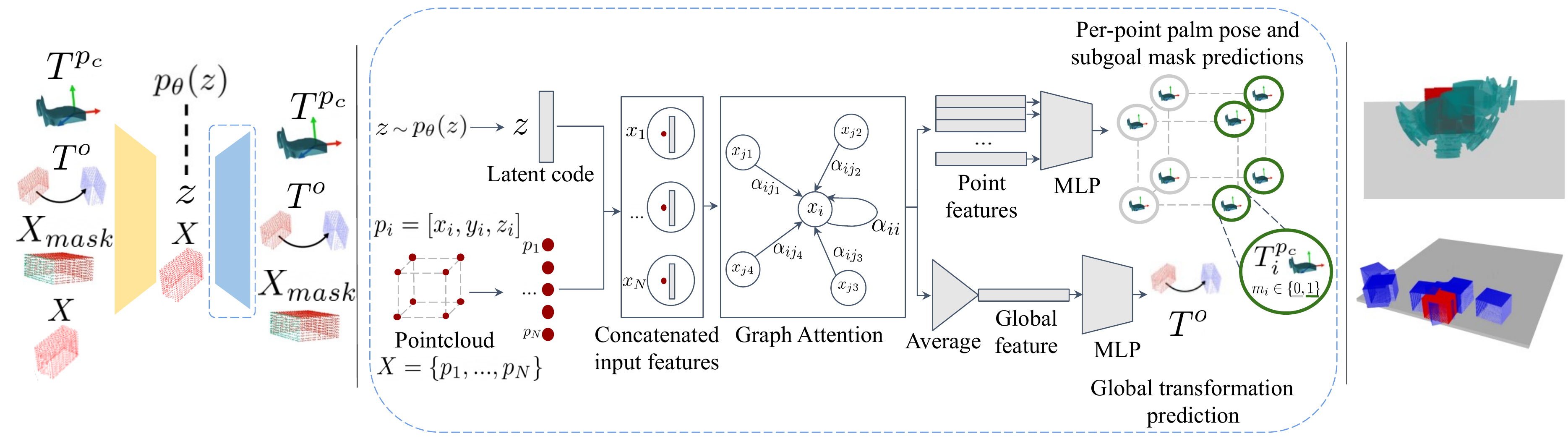}
    \caption{{\bf Left}: CVAE encoder-decoder. The encoder maps the data to a latent distribution which is constrained to match a prior. The decoder takes as input a latent sample $z$ and point-cloud $X$, and outputs the contact pose, object subgoal, and a binary mask, denoted $X_{mask}$, that indicates which part of the object must contact the resting surface. {\bf Middle}: Decoder architecture. Point-cloud coordinates are concatenated with a latent code, and encoded with multiple layers of self-attention into per-point output features and an average-pooled global feature, which are used to predict the contact poses, binary mask, and subgoal transformation. {\bf Right:} Samples of $T^{p_c}$ (top) and $T^o$ (bottom) for \textit{pull} skill}
    \label{fig:network}
    \vspace{-10pt}
\end{figure}

%% file: figText/pipeline.tex
\begin{wrapfigure}{r}{.6\linewidth}
    \vspace{-10pt}
    \centering
    \includegraphics[width=\linewidth]{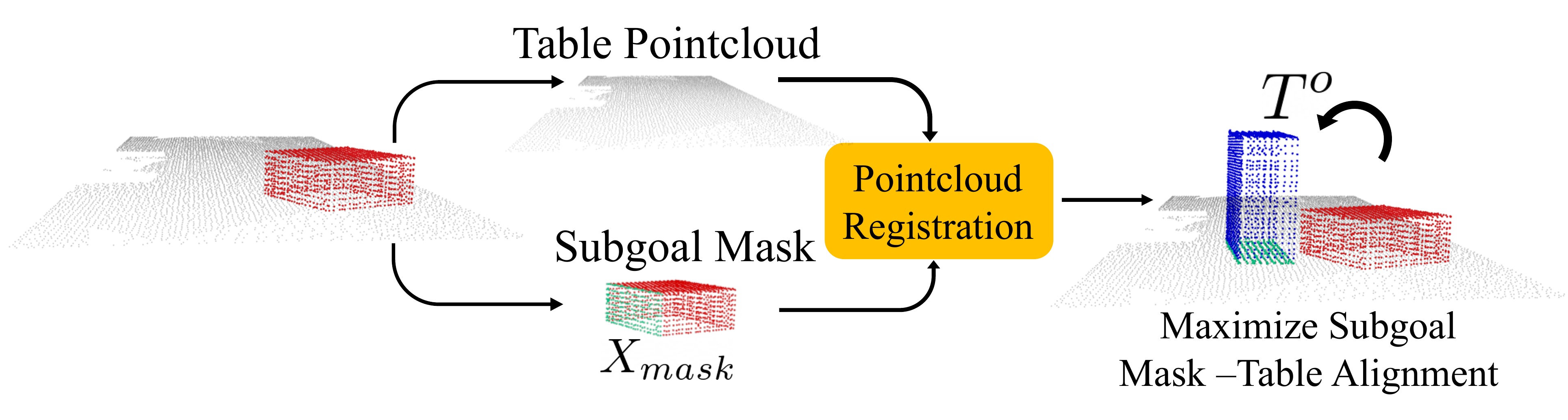}    
    \caption{Illustration of reorientation subgoal prediction. A segmentation mask $X_{mask}$ (green) of points on the object pointcloud that should contact the table after executing the skill is predicted by the neural network. A object transformation $T^o$ is obtained through point-cloud registration of $X_{mask}$ onto the table surface.}
    \label{fig:full-pipelines}
    \vspace{-20pt}
\end{wrapfigure}

%% file: text/experiment.tex
\section{Experiments and Results}
\vspace{-5pt}
\label{sec: experiments}
We perform experiments on a dual-arm ABB YuMi robot with palm end-effectors. We simulate the robot in a table-top environment using the PyBullet~\cite{coumans2016pybullet} simulator and the AIRobot library~\cite{airobot2019}. We place 4 RGB-D cameras with a shared focal point at the table corners, and we use ground truth segmentation masks to obtain segmented point-clouds from simulated depth images. 

\subsection{Performance Evaluation on Single-Step Manipulation Tasks}
\label{sec: single-step-performance-eval}
We first evaluate the efficacy of various design choices used to construct the subgoal and contact pose sampler on the task of single-step manipulation on 19 novel cuboids of different sizes. To generate evaluation data, we initialized each cuboid in 6 uniformly and randomly sampled poses while ensuring that the cuboid rests on the table. Given the point-cloud observation of the object, $X$, we predicted $T^o,T^{p_c}$ from the learned sampler. If the sampler was accurate, then execution of $\pi(T^o,T^{p_c})$ (see Equation~\ref{eq:skills}) should transform the object by $T^o$. After execution, we use the simulator state to evaluate the error between the desired $(T^o)$ and actual object transformation. We define the \textit{success rate} as the percentage of trials where the model finds feasible samples within 15 attempts \textit{and} the executed object transformation is within a threshold of (3 cm/20$^\circ$) of the desired subgoal.

Results in Figure~\ref{fig: single-step-bar} compare the performance of PointNet++~\cite{qi2017pointnet++} against the proposed GAT architecture (see Section~\ref{sec: methods}) used for constructing the CVAE.  It can be seen that GAT (blue) outperforms PointNet++ (orange) on the tasks of grasping and pushing. The performance of both architectures is similar for pulling because it is a much easier manipulation to perform. To gain further insights into differences between the performance of GAT and PointNet++, we visualized the contact poses predicted by both the models in Figure~\ref{fig:gat-grasps} and~\ref{fig:pointnet-grasps} respectively. It can be seen that predictions from PointNet++ are more likely to result in the robot's palms not aligning with the object (see Figure~\ref{fig:pointnet-grasps}). This misalignment results in less stable point contacts at edges instead of patch contacts on faces, which results in inaccurate or infeasible manipulation. Contact with edges also leads to object motion while the robot is trying to grasp, which breaks the sticking-contact assumption and leads to inaccurate prediction of the transformation of the object.

Next, for the \textit{grasp-reorient} skill, we evaluate if predicting the mask-based subgoal (see Section~\ref{sec:mask-representation}) indeed achieves better performance than directly predicting $T^o$. Results in Figure~\ref{fig: single-step-bar-grasp} justify this choice. Qualitative examples of the type of errors made by directly predicting $T^o$ are visualized in Figure~\ref{fig:infeasible-subgoals-predicted-transform}. It can be seen that these examples correspond to physically infeasible transformations, an error-mode which is fixed by using analytical computation of $T^o$ using the mask-based representation. In contrast to \textit{grasp-reorient}, primitives \textit{pull} and \textit{push} only operate in $SE(2)$ and therefore do not suffer from this error mode. For these skills, directly predicting $T^o$ suffices for good performance.

Finally, we also contrast the performance of our method of jointly modeling subgoals and contacts against a baseline that models them independently. Results in Figure~\ref{fig: single-step-bar} shows that joint modeling leads to significant performance gains. 

\input{figText/single_step}
\input{figText/grasp_motion_plan}

\subsection{Performance Evaluation on Multi-step Manipulation Tasks}
\label{sec: multistep-performance-eval}
We now evaluate the performance of our planning algorithm (\sect{sec: multi-step-planning}) when coupled with either our learned skill samplers or with a set of manually designed samplers. The experiment is designed to test the hypothesis that our learned samplers enable efficient multi-step planning with unseen objects for a variety of reconfiguration tasks. The baseline samplers use a combination of heuristics (plane segmentation, antipodal point sampling, and palm alignment with estimated point-cloud normals) based on the privileged knowledge that the objects are cuboids (see Appendix~\ref{sec: app-evaluation-details} for details). 

We created the testing set using 20 novel cuboidal objects. First a plan skeleton is chosen from $\{pg, gp, pgp\}$, where $p$ and $g$ denote \textit{pull} and \textit{grasp-reorient}, respectively. For each skeleton, we construct 200 instances of planning problems. We evaluate the agent's performance using following metrics: (i) planning success rate with a fixed planning budget of 5-minutes, (ii) average final pose error, and (iii) average planning time required over all the trials where a plan is found.  The results in Table~\ref{tab: multistep-results} indicate that our learned models provide a large benefit in planning efficiency over the baseline that was hand-designed for cuboidal objects. The baseline's lower performance is primarily due to the sampling of many infeasible sequences. In contrast, our model learns to bias samples toward parameters that are likely to be feasible and lead to successful execution (see Appendix~\ref{sec: app-additional-discussion} for more detailed failure mode analysis and comparison). The similar pose errors indicate feasible plans obtained by both samplers are of similar quality.

\input{figText/multi_step}

\subsection{Qualitative Results and Capabilities}
\myparagraph{Multiple Placement Surfaces}
\input{figText/bookshelf}
In general, enabling a system to generalize to environments with multiple support surfaces (i.e., a table and a shelf) may be difficult if the system has only been trained using a single support surface. However, our mask-based subgoal representation \textit{directly} enables sampling subgoals on different surfaces. This is because arbitrary target point-clouds, obtained using plane segmentation, can be used during registration. We demonstrate this capability in Figure~\ref{fig: bookshelf}, where a book is moved from the table to a vertical pose on an elevated shelf with a 5-step plan.

\myparagraph{Generalization to Novel Geometry Classes} Figure~\ref{fig:simulated-object-generalization-single} shows \textit{grasp-reorient} predictions trained only on cuboids, and tested on a cylindrical object and on an object in the shape of a bottle. Predictions for both $T^o$ and $T^{p_c}$ are quite sensible and were executed successfully in the simulator. Figure~\ref{fig:simulated-object-generalization-multi} shows a 3-step plan found with our planning algorithm on the bottle-shaped object. Appendix~\ref{sec: app-additional-results} and \ref{sec: app-additional-discussion} contains additional results and discussions related to generalization to novel geometry. 
\input{figText/object_generalization_sim}

\myparagraph{Real Robot Experiments} 
\input{figText/real_robot}
In light of COVID-19, access to robots in our lab has been limited and we focus only on qualitatively demonstrating the capabilities of our framework on a real ABB YuMi robot.
%
Figure~\ref{fig:real-robot-single} shows snapshots from skills being executed on a variety of objects after training only in simulation.  
%
Figure~\ref{fig:real-robot-multi} shows a 2-step plan obtained by the full framework being executed with a bottle. We used calibrated Intel RealSense D415 RGB-D cameras at the table corners to obtain point-clouds, AR tags to specify $T^o_{des}$ for multi-step planning, and a pre-trained neural network \cite{he2017mask, wu2019detectron2} combined with point-cloud cropping heuristics for segmentation (details in Appendix~\ref{sec: app-implementation-details}). 
Finally, to quantitatively evaluate performance on realistic perceptual inputs, we conducted additional simulated experiments (as in Section~\ref{sec: multistep-performance-eval}) using noise-corrupted point-clouds. The results showed a negligible performance change, further validating real-world applicability of our system (see Appendix~\ref{sec: app-additional-results} for details).

%% file: figText/single_step.tex
\begin{figure}[t]
    \centering
    \begin{tabular}{ccc}
    \begin{subfigure}[b]{0.27\linewidth}
		\includegraphics[width=1.0\linewidth]{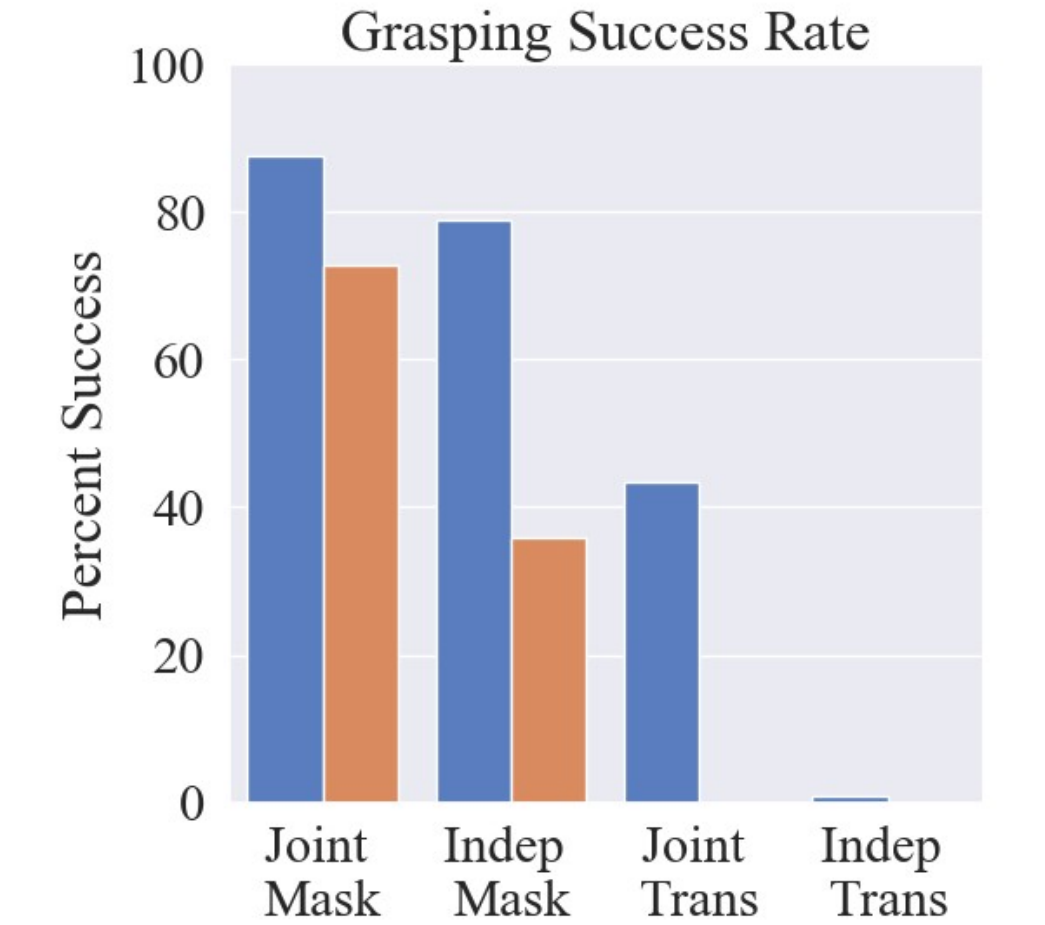}
		\caption{} 
		\label{fig: single-step-bar-grasp}
	\end{subfigure}&
    \begin{subfigure}[b]{0.27\linewidth}
		\includegraphics[width=1.0\linewidth]{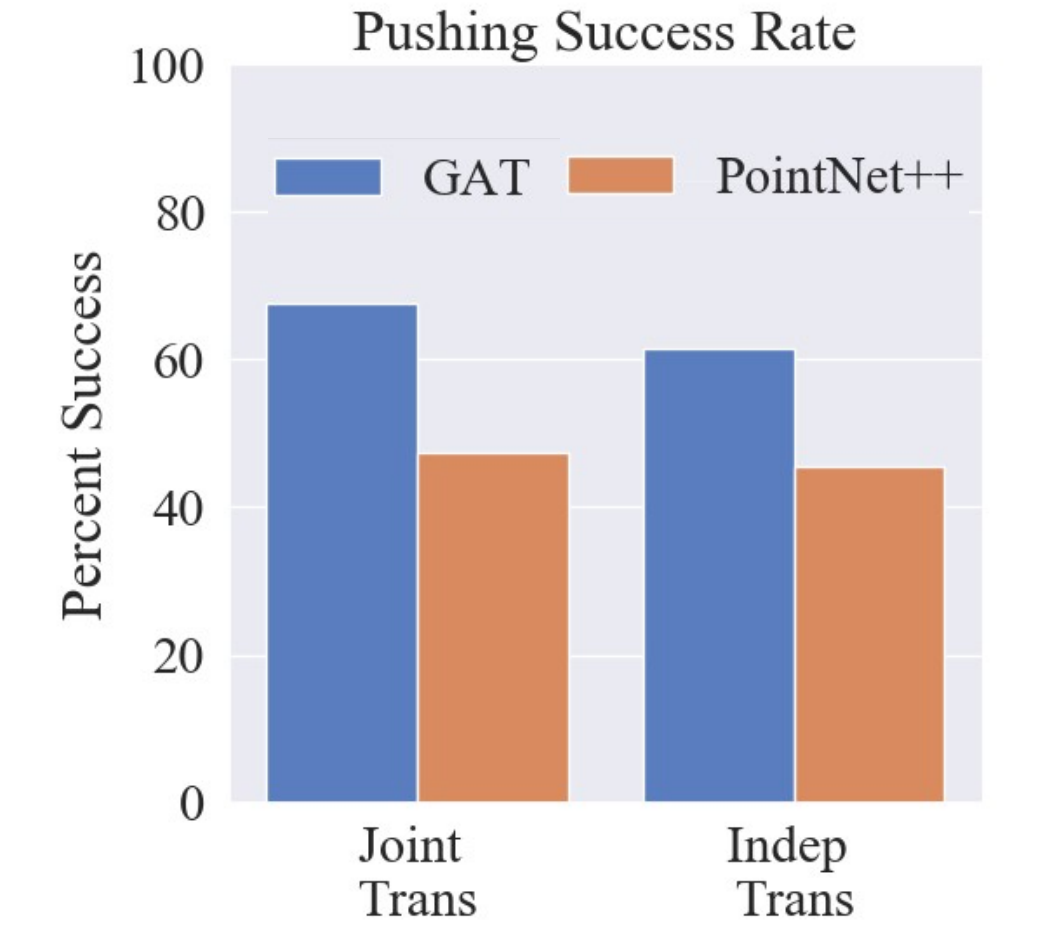}
		\caption{} 
		\label{fig: single-step-bar-push}
	\end{subfigure}&	
    \begin{subfigure}[b]{0.27\linewidth}
		\includegraphics[width=1.0\linewidth]{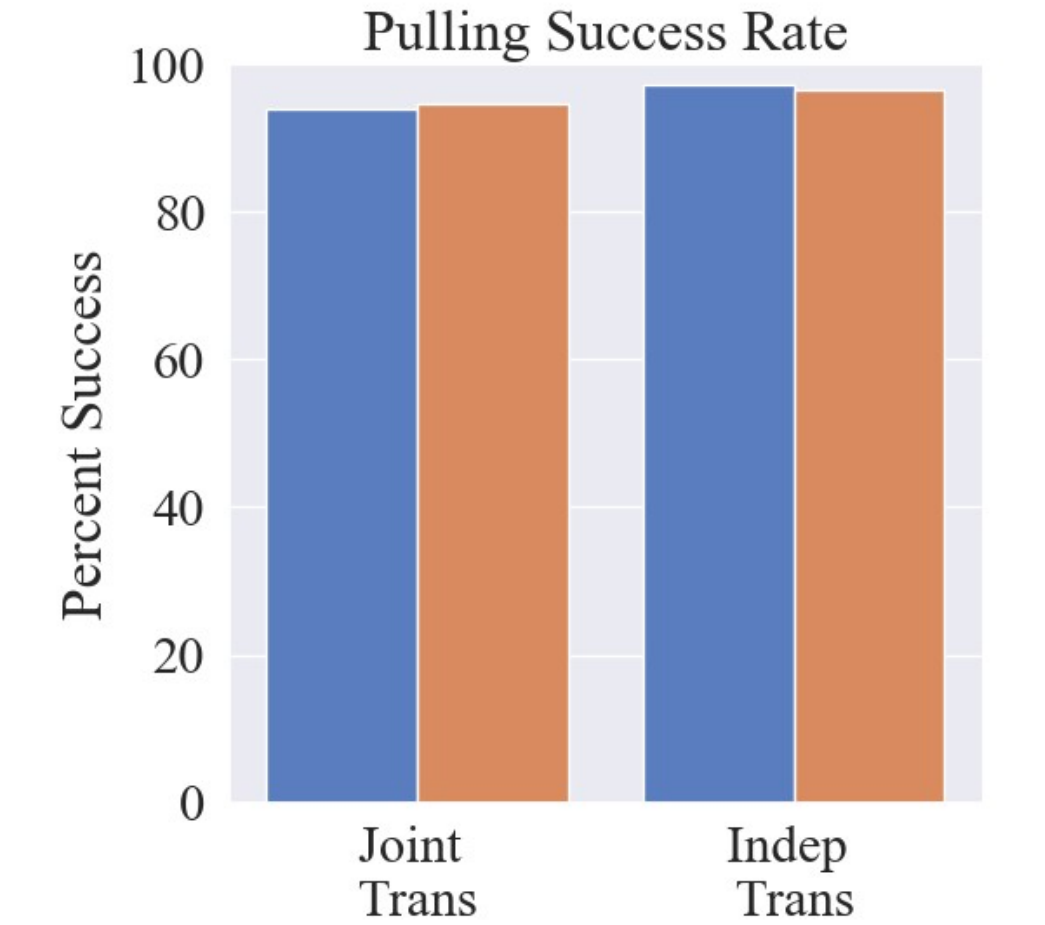}
		\caption{}
		\label{fig: single-step-bar-pull}
    \end{subfigure}
    \end{tabular} 
    \vspace{-5pt}
    \caption{Success rates for single step grasping, pushing, and pulling. Joint Mask refers to using $X_{mask}$ and predicting $T^o$ and $T^{p_c}$ together, Indep Mask refers to using $X_{mask}$ but predicting $T^o$ and $T^{p_c}$ separately, and Joint Trans refers to not using $X_{mask}$ and predicting $T^o$ and $T^{p_c}$ together. The overall results indicate a large value provided by using our point-cloud encoder, mask subgoal prediction, and joint prediction scheme. See Appendix~\ref{sec: app-additional-results} for more detailed results breakdown.}
    \label{fig: single-step-bar}
\end{figure}

%% file: figText/grasp_motion_plan.tex
\begin{figure}
    \centering
  \begin{tabular}{c|c}
     \begin{subfigure}{0.45\linewidth}
		\includegraphics[width=1.0\linewidth]{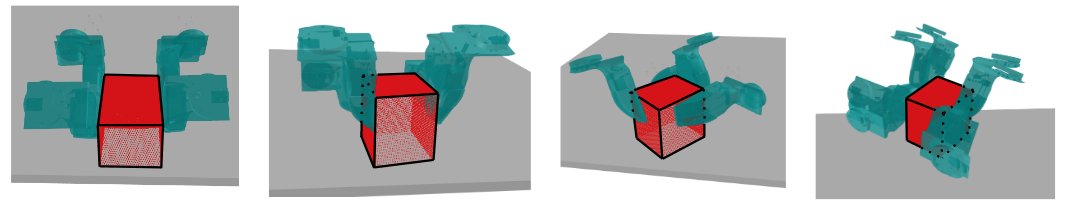}
		\caption{Predicted grasps from GAT (ours).} 
		\label{fig:gat-grasps}
	\end{subfigure}\\
    \begin{subfigure}{0.45\linewidth}
		\includegraphics[width=1.0\linewidth]{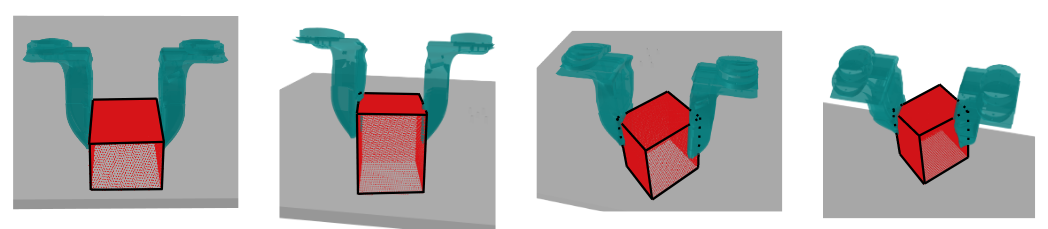}
		\caption{Predicted grasps from PointNet++.}
		\label{fig:pointnet-grasps}
    \end{subfigure}&
    \multirow[t]{2}{*}[-22pt]{\begin{subfigure}[b]{0.4\linewidth}
		\includegraphics[width=1.0\linewidth]{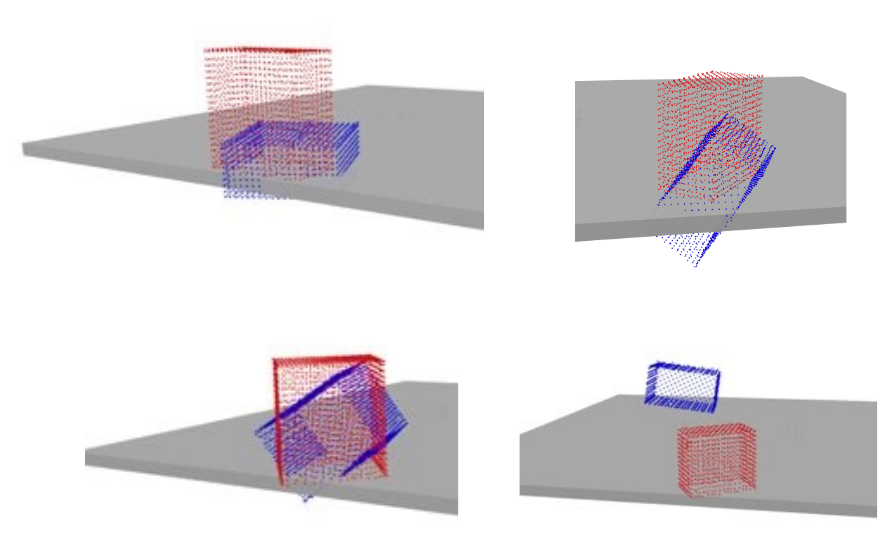}
		\caption{Result of directly predicting $T^o$.}
		\label{fig:infeasible-subgoals-predicted-transform}
    \end{subfigure}}
    \end{tabular}    
    \caption{Visualization of predicted grasps from GAT and PointNet++ are presented in rows (a) and (b) respectively. (c) Using the skill sampler to directly predict $T^o$ leads to inaccurate predictions, because small errors in $T^o$ can correspond to physically infeasible configurations (blue point-clouds). Our proposed approach using a mask-based sub-goal representation overcomes this issue. }
    \label{fig: gat-pointnet-grasp-samples}
    \vspace{-10pt}
\end{figure}

%% file: figText/multi_step.tex
\def\arraystretch{1.0}
\begin{table}[t]
\small
\centering
\caption{Comparing the multistep planning success rate, average planning time, and error in achieving the target configuration between learned and hand-designed samplers.  Values were obtained over 200 trials, with error bars denoting a 95\% confidence interval. \texttt{P}: \texttt{PullRight}, \texttt{G}: \texttt{GraspReorient}}. 
\begin{tabular}{l|cccc}
\hline \hline
Type & \multirow{1}{*} Skeleton & Success Rate (\%) & Pose Error (cm / deg) & Time (s) \\ \hline 
 \multirow{3}{*}{Learned} & \texttt{P}$\shortrightarrow$\texttt{G} & 96.9 $\pm$ 2.3 & 0.7 $\pm$ 0.5 ~/~ 3.0 $\pm$ 1.3 & 36.0 $\pm$ 6.9 \\
 & \texttt{G}$\shortrightarrow$\texttt{P} & 95.9 $\pm$ 2.8 & 3.8 $\pm$ 0.4 ~/~ 28.2 $\pm$ 4.9 & 30.5 $\pm$ 7.1\\
 & \texttt{P}$\shortrightarrow$\texttt{G}$\shortrightarrow$\texttt{P} & 86.9 $\pm$ 4.8 & 2.2 $\pm$ 0.4 ~/~ 18.9 $\pm$ 3.9 & 63.7 $\pm$ 10.6\\
\hline
\multirow{3}{*}{Hand Designed} & \texttt{P}$\shortrightarrow$\texttt{G} & 32.2 $\pm$ 6.5 & 0.9 $\pm$ 0.4 ~/~ 8.3 $\pm$ 5.0 & 110.0 $\pm$ 19.8\\
 & \texttt{G}$\shortrightarrow$\texttt{P} & 70.4 $\pm$ 6.3 & 3.2 $\pm$ 0.5 ~/~ 18.1 $\pm$ 5.2 & 65.4 $\pm$ 12.9 \\
 & \texttt{P}$\shortrightarrow$\texttt{G}$\shortrightarrow$\texttt{P} & 54.3 $\pm$ 7.0 & 2.4 $\pm$ 0.8 ~/~ 16.2 $\pm$ 6.5 & 69.7 $\pm$ 15.1 \\
\hline \hline
\end{tabular}
\label{tab: multistep-results}
\end{table}

%% file: figText/bookshelf.tex
\begin{figure}[t!]
    \vspace{-10pt}
    \centering
    \includegraphics[width=\linewidth]{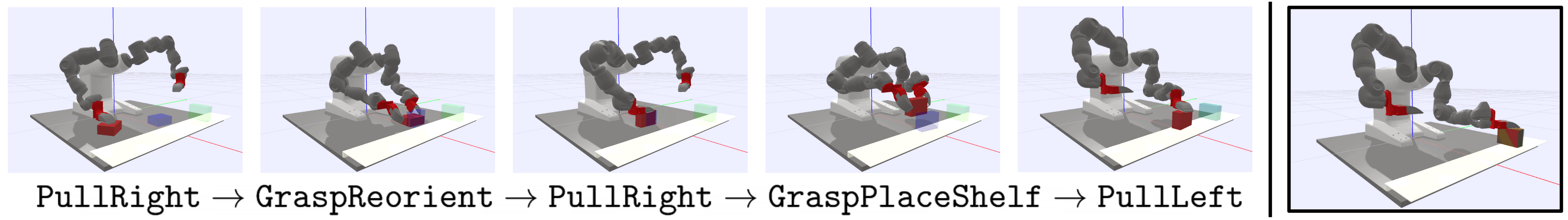}  
    \caption{We show execution of a 5-step plan skeleton, where the goal configuration is on an elevated shelf. Sampling a subgoal on a shelf fits directly in our framework, through the use of our segmentation mask-based reorientation subgoal representation.}
    \label{fig: bookshelf}
    \vspace{-15pt}
\end{figure}

%% file: figText/object_generalization_sim.tex
\begin{figure}
    \centering
  \begin{tabular}{cc}
    \begin{subfigure}[b]{0.31\linewidth}
		\includegraphics[width=1.0\linewidth]{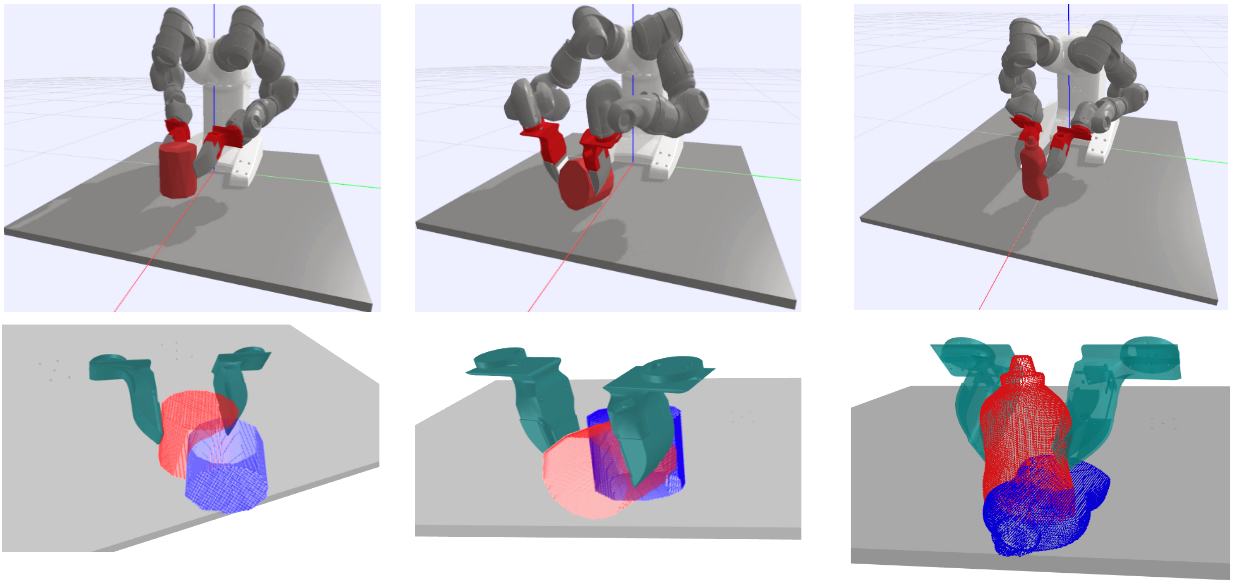}
		\caption{} 
		\label{fig:simulated-object-generalization-single}
	\end{subfigure}\hfill&
    \begin{subfigure}[b]{0.6\linewidth}
		\includegraphics[width=1.0\linewidth]{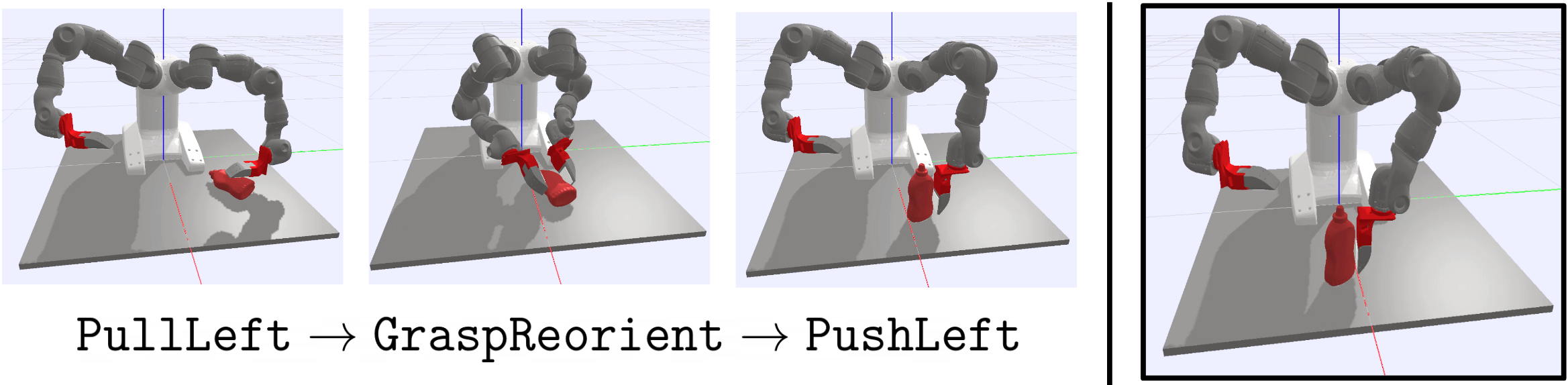}
		\caption{}
		\label{fig:simulated-object-generalization-multi}
    \end{subfigure}\hfill
    \end{tabular}    
    \vspace{-5pt}
    \caption{(a) Predictions made using point-clouds from novel geometric classes, by our model which was only trained on cuboids. (b) Executing a plan for a 3-step plan skeleton on a bottle-shaped object}
    \label{fig: object-generalization-simulated}
    \vspace{-10pt}
\end{figure}

%% file: figText/real_robot.tex
\begin{figure}
    \centering
  \begin{tabular}{cc}
    \begin{subfigure}[b]{0.5\linewidth}
		\includegraphics[width=1.0\linewidth]{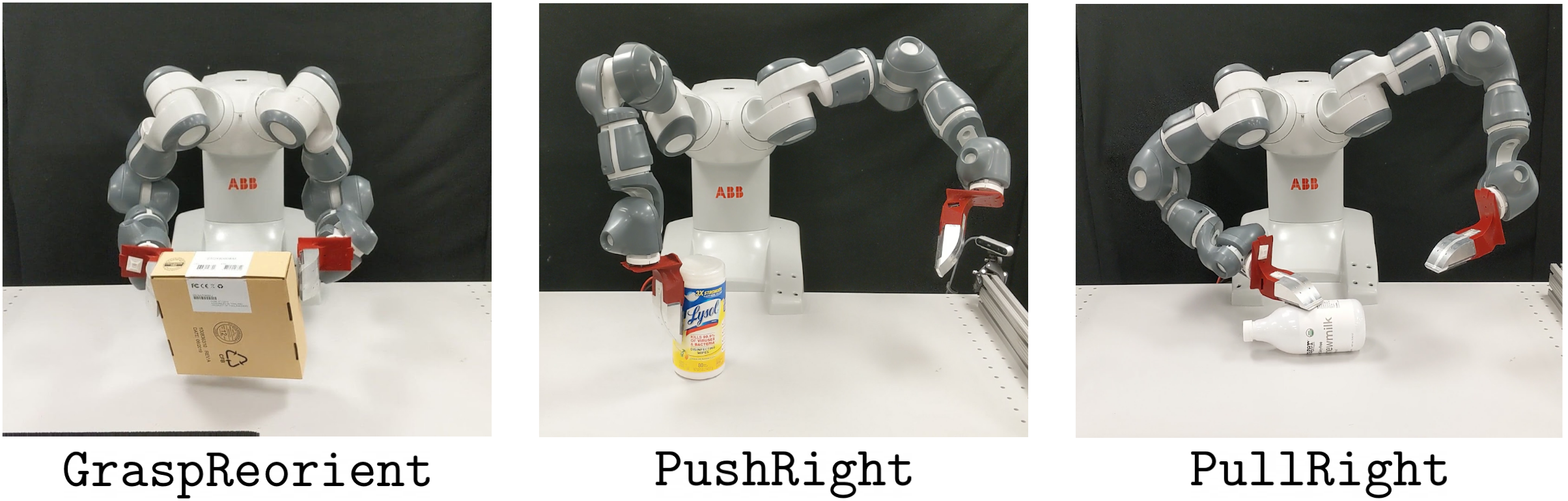}
		\caption{} 
		\label{fig:real-robot-single}
	\end{subfigure}\hfill&
    \begin{subfigure}[b]{0.39\linewidth}
		\includegraphics[width=1.0\linewidth]{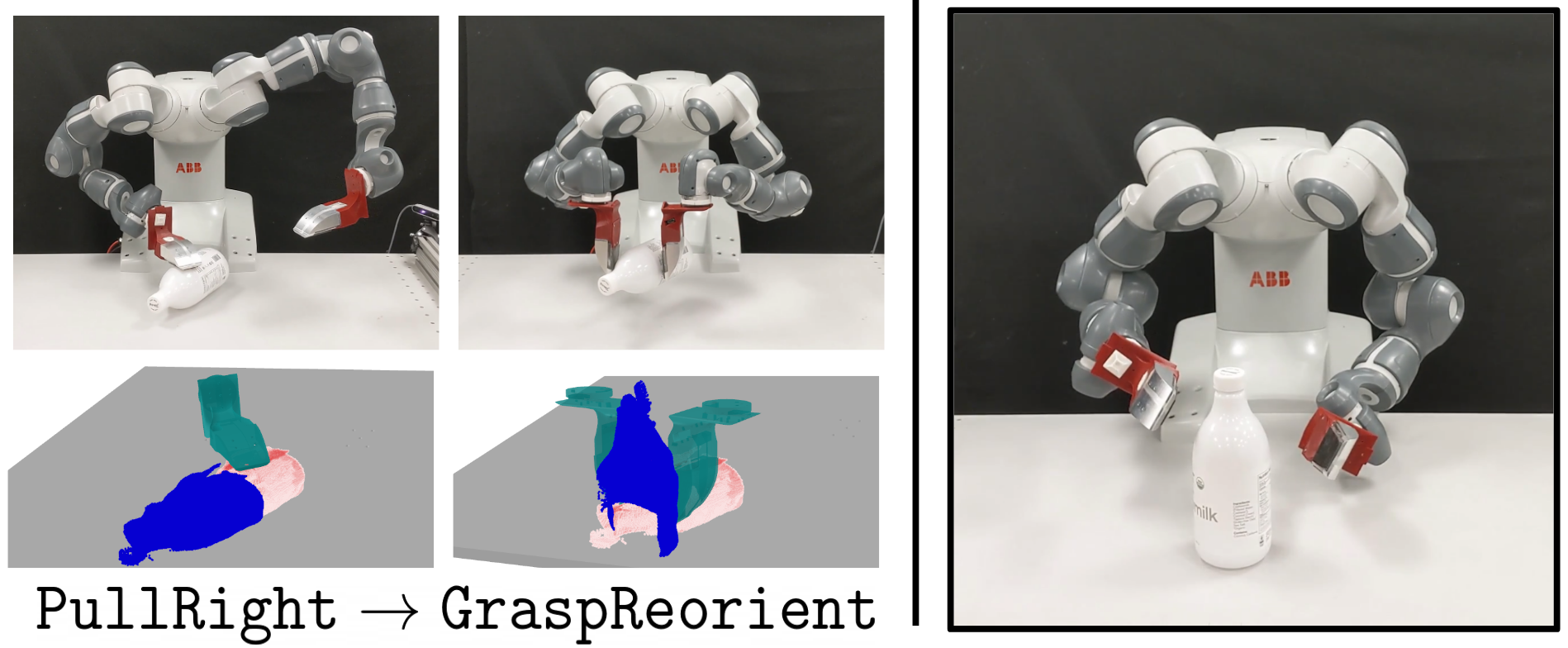}
		\caption{}
		\label{fig:real-robot-multi}
    \end{subfigure}\hfill
    \end{tabular}    
    \vspace{-5pt}
    \caption{(a) Execution of skills using parameters obtained from our learned samplers with real point-clouds as input. (b) 2-step plan executed on a real bottle-shaped object.}
    \label{fig: real-robot}
    \vspace{-15pt}
\end{figure}

%% file: text/related_work.tex
\vspace{-5pt}
\section{Related Work}
\label{sec: related}
\myparagraph{Perception-based Multi-step Manipulation}
\label{sec: afford-reason}
Most similar to this work is \citet{fang2019cavin}, which decouples subgoals and robot actions and learns samplers to enable multistep planning from segmented object point-clouds. In contrast to us, they learn neural network samplers for forward simulation both in the subgoal and the action space. We bypass the problem of learning low-level dynamics in the action space and use $SE(3)$ transformations to forward propagate subgoals. Moreover, they only considered planar tasks with a straight-line push primitive. In \citep{zeng2018vpg}, a pixel-wise Q-function represents high quality contact locations, and reinforcement learning is used to obtain sequencing behavior between pushing and grasping.

\myparagraph{Task-Oriented Contact-Affordances}
\label{sec: tog}
In \citep{fang2020tog}, a DexNet~\citep{mahler2017dex} grasp detector is extended to use a task-specific reward to select grasps that are useful for a downstream task. In \citep{manuelli2019kpam}, keypoint affordances are used for task specification, grasping, and planning motions to manipulate categories of objects. \citep{qin2019keto} extends this approach by learning the keypoints based on the task rather than manually specifying them. \citep{kloss2019accurate} combines learned push affordances with a mechanics model to rank push locations based on predicted outcomes and the task at hand. We approach predicting task-oriented contacts by modeling the joint distribution over reachable subgoals and corresponding contacts for a set of primitive skills.  

\myparagraph{Manipulation with 3D Deep Learning}
\label{sec: 3d}
\citet{mousavian2019graspnet} and \citet{murali2019collisionnet} train a VAE~\citep{kingma2013auto} based on PointNet++~\citep{qi2017pointnet++} to generate and refine feasible grasps given partial point-clouds. In \cite{byravan2017se3}, the authors propose a scene dynamics model trained to predict $SE(3)$ transformations of point-clouds, but their model operates over low-level actions rather than high-level skills, limiting the object manipulation demonstrations to short-horizon interactions.

\myparagraph{Learning to guide TAMP}
Several algorithms exist for learning to guide TAMP planners~\cite{beomjoon2017icra,beomjoon2018sampler,beomjoon2019experience,chitnis2019, chitnis2016guided}. In \cite{beomjoon2017icra}, the authors consider a similar problem setup as ours in which the plan skeleton is given, and the goal is to find the continuous parameters of the skeleton. Instead of learning samplers for each manipulation skill, they learn to predict the parameters for the entire skeleton. \cite{beomjoon2018sampler,beomjoon2019experience, chitnis2019}, like us, consider learning a sampler for each manipulation skill to improve planning efficiency. Unlike our setup, however, these methods assume that the poses and shapes of objects are perfectly estimated. Our framework can be seen as extending these works to problems in which the robot must reason about objects with unknown shapes and poses by directly learning to map a sensory observation to a distribution over promising parameters of manipulation skills.

%% file: text/conclusion.tex
\vspace{-5pt}
\section{Conclusion}
\label{sec:conclusion}
This paper presents a method to enable multistep sampling-based planning using primitive manipulation skills when provided with segmented object point-clouds. Our approach uses deep conditional generative modeling to map point-cloud observations to a distribution over likely contact poses and subgoals, which are used as inputs to the skills. Novel technical aspects of our approach enable the ability to use the skills effectively, and our learned samplers provide planning efficiency gains over a manually designed baseline sampler. Our qualitative results validate that the approach transfers to real point-clouds, scales toward manipulating more complex object geometry, and enables solving more sophisticated multistep tasks.

%% file: text/acknowledgement.tex

\acknowledgments{This work was supported in part by the DARPA Machine Common Sense Grant and the Amazon Research Awards. Anthony and Yilun are supported in part by NSF Graduate Research Fellowships. We would like to thank the anonymous reviewers for their helpful comments and feedback.}

%% file: text/appendix.tex
\appendix
\renewcommand{\thesection}{A.\arabic{section}}
\renewcommand{\thefigure}{A\arabic{figure}}
\setcounter{section}{0}
\setcounter{figure}{0}

\pagebreak
\section{Appendix}
\label{sec: appendix}

\subsection{Point-cloud Registration}
\label{sec: app-point-cloud-registration-icp}
For the mask-based representation, we use Iterative Closest Point (ICP) \cite{paul1992beslICP} to solve for the registration and obtain the subgoal transformation. The segmented table point-cloud is used as the target. The initial transformation is combined from two sources: (1) the planar translational component of the neural network's direct subgoal prediction, and (2) a pure forward $\frac{\pi}{2}$ pitch about the object body frame. We use the Open3D~\cite{Zhou2018open3d} implementation of point-to-point ICP, which we find to work effectively. Note that other registration techniques and methods for obtaining an initial transformation could equivalently be used in our framework.

Subgoal predictions for the \textit{pull} and \textit{push} skills, which only operate in $SE(2)$, do not suffer from the same physical inaccuracy issues as \textit{grasp-reorient}. Therefore, subgoals for these skills are sampled by projecting $T^o$ predicted from the model to $SE(2)$.

\subsection{Sampling-based Planning Algorithm}
\label{sec: app-multistep-planning-algo}

Algorithm \ref{alg: mutlstep_planning} describes our sampling-based planning algorithm in detail. Nodes in a search tree are initialized with InitPointCloudNode. Assuming some initial point-cloud and samples for the skill subgoal and contact pose, a node is specified based on the subgoal and contact pose, the resulting point-cloud after being transformed by the subgoal transformation, and the parent node whose point-cloud was used to sample the corresponding parameters. 

Buffers for each step in the plan skeleton are used to store nodes containing parameters that are found to be feasible. When sampling at each step, a node is randomly popped from the buffer for the corresponding step $t$, and the point-cloud from that node is used as input for the skill sampler. Parent nodes are specified based on their index in their respective buffer.

The SatisfiesPreconditions procedure operates on the point-cloud from the popped node to determine if further subgoal/contact sampling will proceed. These precondition checks improve efficiency by avoiding sampling many point-cloud configurations that are likely or guaranteed to fail by leveraging reachability constraints that are easy to check and specify with respect to the point-cloud. For instance, for \textit{grasp-reorient}, we only accept point-clouds whose average planar positional coordinates are within a valid rectangular region near the front-center of the table. A similar precondition for \textit{pull} and \textit{push} checks if the average planar point-cloud position is within the boundaries of the table. We use the same preconditions sampler variants.

The FeasibleMotion procedure checks if the motion corresponding to the sampled parameters is feasible. This process includes two subprocedures: (1) checking for collisions between the robot and the environment at the initial and final configurations of the skill, (2) checking for collisions and kinematic feasibility of the motion computed by the low-level primitive planners. Both subprocedures must pass for the skill to be feasible. We only run subprocedure (2) if subprocedure (1) passes, as (1) is computationally cheaper and helps filter out infeasible motions more efficiently. At each step in $PS$, the skill samplers have a maximum number of samples, $K_{max}$, to obtain a set of feasible parameters. If they fail to do so within this limit, sampling moves on to the next step in the skeleton. On the final step $T$, the required transformation is directly computed based on the sequence of subgoal transformations that leads to the node that has been popped, starting from the root node. If the final subgoal transformation is not feasible, sampling begins all over again at the first step in $PS$, so that new start point-clouds continue to be added to the buffers. In our experiments we used $K_{max} = 10$.

When a feasible set of parameters is found on the final step, the ReturnPlan procedure is used to backtrack from the final node to the root node along the parents (similarly as in RRT \cite{lavalle1998rapidly}) and return the plan to be executed. If a feasible plan is not found within 5 minutes, the planner returns a failure.

\input{algText/planning_alg}


\subsection{Conditional Variational Auto-encoder Skill Samplers}
\label{sec: app-cvae-sampler}
The CVAE is trained to maximize the evidence lower bound (ELBO) of the conditional log-likelihood of the data in $D_{\pi}$. We present a brief description of using the CVAE framework in our setup (for detailed derivation, please see~\cite{kingma2013auto, sohn2015cvae}).

Following \cite{sohn2015cvae}, we denote \textit{inputs} $\mathbf{x}$, \textit{outputs} $\mathbf{y}$, and latent variables $\mathbf{z}$. We wish to model the distribution $p_{\theta}(\mathbf{y} | \mathbf{x})$ (this is what is used as the skill sampler during planning). We can do so by considering a latent variable model, $p_{\theta}(\mathbf{y} | \mathbf{x}) = \int p_{\theta}(\mathbf{y} | \mathbf{x}, \mathbf{z})p_{\theta}(\mathbf{z})d\mathbf{z}$. In doing so we assume our data is generated via the following generative process: given observation $\mathbf{x}$, latent variable $\mathbf{z}$ is drawn from prior distribution $p_{\theta}(\mathbf{z} | \mathbf{x})$ and output $\mathbf{y}$ is generated from the conditional distribution $p_{\theta}(\mathbf{y} | \mathbf{x}, \mathbf{z})$. The training procedure for the CVAE uses neural networks to approximate the generative distribution $p_{\theta}(\mathbf{y} | \mathbf{x}, \mathbf{z})$ along with a recognition model $q_{\phi}(\mathbf{z} | \mathbf{x}, \mathbf{y})$. $q_{\phi}(\mathbf{z} | \mathbf{x}, \mathbf{y})$ and $p_{\theta}(\mathbf{y} | \mathbf{x}, \mathbf{z})$ can be interpreted as a probabilistic encoder and decoder, respectively. The ELBO of the conditional log-likelihood,
\begin{align}
    \log p_{\theta}(\mathbf{y} | \mathbf{x}) \geq -D_{KL}(q_{\phi}(\mathbf{z} | \mathbf{x}, \mathbf{y}) || p_{\theta}(\mathbf{z})) + \mathbb{E}_{q_{\phi}(\mathbf{z} | \mathbf{x}, \mathbf{y})}[\log p_{\theta}(\mathbf{y} | \mathbf{x}, \mathbf{z})]
\end{align}
is used as a surrogate objective function to optimize with respect to parameters $\theta$ and $\phi$. In our setup we assume $p_{\theta}(\mathbf{z}) = p_{\theta}(\mathbf{z} | \mathbf{x})$.

To optimize the ELBO, the encoder $q_{\phi}(\mathbf{z} | \mathbf{x}, \mathbf{y})$ is trained to map the data to a latent conditional distribution, which is constrained by a KL-divergence loss to resemble a prior $p_{\theta}(\mathbf{z})$ (in our case, a unit Gaussian), while the decoder $p_{\theta}(\mathbf{y} | \mathbf{x}, \mathbf{z})$ is trained to use samples from the latent space to reconstruct the data.  An estimator of the ELBO is used as the loss to optimize when training the neural networks. The estimator enables taking gradients with respect to the network parameters, and it is obtained using the reparameterization trick, 
\begin{align}
    \mathcal{L}_{\text{CVAE}}(\mathbf{x}, \mathbf{y}; \theta, \phi) &= -D_{KL}(q_{\phi}(\mathbf{z} | \mathbf{x}, \mathbf{y}) || p_{\theta}(\mathbf{z})) + \frac{1}{L} \sum_{l=1}^{L}\log p_{\theta}(\mathbf{y}|\mathbf{x}, \mathbf{z}^{(l)}) \\
    \mathbf{z}^{(l)} &= g_{\phi}(\mathbf{x}, \mathbf{y}, \epsilon^{(l)}), ~ \epsilon^{(l)} \sim \mathcal{N}(\mathbf{0}, \mathbf{I})
\end{align}

We can use the closed-form expression for the KL divergence between the prior $p_{\theta}(\mathbf{z})$ and the approximate posterior $q_{\phi}(\mathbf{z} | \mathbf{x}, \mathbf{y})$ when both distributions are assumed to be Gaussian~\cite{kingma2013auto}. The expression uses the mean $\mu$ and standard deviation $\sigma$ of the approximate posterior, where $\mu$ and $\sigma$ are outputs predicted by the encoder network.

\textbf{Joint Skill Sampler Models:} In our joint models, observations $\mathbf{x}$ are the point-cloud observation $X$, and output variables $\mathbf{y}$ include $T^o, T^{p_c}$, and point-cloud segmentation mask $X_{mask}$. The latent variable model in this setup is
\[
p(T^o, T^{p_c}, X_{mask} | X) = \int p(T^o, T^{p_c}, X_{mask} | X, \mathbf{z})p(\mathbf{z})d\mathbf{z} 
\]

\textbf{Independent Skill Sampler Models:} In the baselines that use independent models, output variables $\mathbf{y}$ are separated into two independent sets, with corresponding independent latent variables. $\mathbf{y_1}$ includes $T^o$ and $X_{mask}$, while $\mathbf{y_2}$ includes $T^{p_c}$. We represent these as two separate CVAEs:

\begin{align}
    p(T^o, T^{p_c}, X_{mask} | X) &= p(T^o, X_{mask} | X)p(T^{p_c} | X) \\
    p(T^o, X_{mask} | X) &= \int p(T^o, X_{mask} | X, \mathbf{z_1})p(\mathbf{z_1})d\mathbf{z_1} \\
    p(T^{p_c} | X) &= \int p(T^{p_c} | X, \mathbf{z_2})p(\mathbf{z_2})d\mathbf{z_2}
\end{align}
%


\subsection{Neural Network Architecture Details}
\label{sec: app-model-architecture-details}
\input{figText/gat_architectures}
\input{figText/pointnet_architectures}

\textbf{Inputs and Outputs}
Subgoals $T^o$ and contact poses $T^{p_c}$ are represented as 7-dimensional vectors, made up of a 3D position for the translational component and a unit quaternion for the rotational component. Unit quaternions are directly regressed as an unnormalized 4-dimensional vector, which is normalized after prediction.

During our experiments, we found overall training and evaluation performance is improved when the decoder is trained to predict $X_{mask}$ and $T^o$ together (i.e., as complementary auxiliary tasks), even if $T^o$ is not used for executing the skill. Based on this observation and to simplify implementation, all decoder models that are used to predict subgoals were set up in this way to predict $T^o$ and $X_{mask}$ jointly.

Before being input to any of the neural networks, the point-cloud $X$,
\[X = \begin{bmatrix}\mathbf{p_1} \\ ... \\ \mathbf{p_N} \end{bmatrix} = \begin{bmatrix}x_i, y_i, z_i \\ ... \\ x_N, y_N, z_N\end{bmatrix}\]
is pre-processed with the following steps: $X$ is converted into $\bar{X}$, by computing the mean over all the points $X_c = \sum_{i=1}^{N} \mathbf{p_i}$ = $[x_c, y_c, z_c]$, subtracting $X_c$ from all the points in $X$, and concatenating $X_c$ as an additional feature to all the zero-mean points to obtain 
\[\bar{X} \in \mathbb{R}^{N \times 6} = \begin{bmatrix} \mathbf{p_1} - X_c \oplus X_c \\ ... \\ \mathbf{p_N} - X_c \oplus X_c \end{bmatrix}\]

We uniformly downsample the observed point-cloud to a fixed number of points before converting to $\bar{X}$ and computing the network forward pass. We use $N=100$ points in all experiments with learned models. For the baseline samplers, we use the full dense point-cloud, which can have a variable number of points.

\textbf{Model Architecture based on Graph Attention Networks}
Figure \ref{fig: gat-architecture-detailed} shows the GAT-based architectures for the CVAE encoder and decoder. The encoder is a Graph Attention Network, which is trained to represent the approximate posterior $q_{\phi}(\mathbf{z} | \mathbf{x}, \mathbf{y})$. To combine the input variables $\mathbf{x}$ ($\bar{X}$), and the output variables $\mathbf{y}$, we concatenate all the variables in $\mathbf{y}$ as extra point-wise features to $\bar{X}$. $T^o$ and $T^{p_c}$ are repeated and concatenated with all the points in $\bar{X}$, and the point-wise binary features in $X_{mask} \in \mathbb{R}^{N}$ are directly concatenated with the corresponding points in $\bar{X}$.

Output point features are computed using the graph-attention layers in the encoder. A mean is taken across all the points to obtain a global feature encoding, which is then provided to separate fully-connected output heads that compute $\mu$ and $\sigma$. $\mu$ and $\sigma$ are reparameterized to compute $\mathbf{z}$, and are used to compute the KL divergence loss~\cite{kingma2013auto} in $\mathcal{L}_{\text{CVAE}}$.

The decoder receives as input $\bar{X}$ and $\mathbf{z}$. $\bar{X}$ is projected to a higher dimensional space with a learnable weight matrix $W$. $\mathbf{z}$ is repeated and concatenated with each point in $W\bar{X}$ as an additional per-point feature. This representation is then provided to another Graph Attention Network to obtain output point features. 

The point features are used by different output heads to reconstruct the data. \textit{Each} point feature is independently used by a fully-connected output head to predict the contact pose and another fully-connected output head followed by a sigmoid to predict the binary mask. In parallel, the point features are averaged and passed to a third fully-connected head that predicts $T^o$.

\textbf{Model Architecture based on PointNet++}
The encoder and decoder have a PointNet++ architecture~\cite{qi2017pointnet++}, shown in Figure \ref{fig: pointnet-architecture-detailed}. $X$ is converted to $\bar{X}$ and concatenated with all the $\mathbf{y}$ variables before being passed to the encoder, in the same way as described above . The PointNet++ encoder computes a global point-cloud feature, and fully-connected output heads predict $\mu$ and $\sigma$ which are reparameterized as $\mathbf{z}$.

During decoding, $X$ converted to $\bar{X}$, $\mathbf{z}$ is concatenated as a per-point feature with $\bar{X}$, and a global point-cloud feature is computed with a PointNet++ decoder. Separate fully-connected heads use the global feature to make a single contact pose prediction for $T^{p_c}$ and transformation $T^o$. The global feature is concatenated to all the original points in $\bar{X}$ and passed to a third fully-connected output that predicts binary point labels for $X_{mask}$.

\textbf{CVAE Inference} 
During testing, a latent vector $\mathbf{z}$ is sampled from the prior $p_{\theta}(\mathbf{z})$ and the trained decoder follows the same respective processes as described above with the trained [GAT or PointNet++] layers and output heads. We find it worked well to randomly sample one of the per-point $T^{p_c}$ predictions for use in the executed skill when using the GAT-based model. In our experiments, we used latent vectors of dimension 256 for the joint models and 128 for independent contact and subgoal models.


\subsection{Training Data Generation and Training Details}
We generate training data by simulating the primitive skills with the known 3D models of the cuboids, sampling transitions between stable object poses for subgoals, and sampling the object mesh for contact poses. If the skill is feasible and successfully moves the object to the subgoal, the parameters are added to the dataset. We also record the points in the point-cloud that end within Euclidean distance thresholded from the table as ground truth labels for $X_{mask}$.

The number of objects and dataset size used for training each skill sampler are shown in Table \ref{tab: training-data-statistics}. Because the data generation process has a high degree of stochasticity and the amount of time required to obtain a large number of successful samples varies between skills, the number of data points/objects used for each skill was not consistent. 

\input{figText/training_table}

During training, we use the Adam optimizer~\cite{kingma2014adam} with learning rate 0.001 to minimize the $\mathcal{L}_{\text{CVAE}}$ loss function with respect to the parameters of the CVAE encoder and decoder. The reconstruction term in the loss is broken up into separate components: We use an MSE loss for the positional components of $T^{o}$ and $T^{p_c}$, and the below geodesic orientation loss~\cite{kuffner2004effective, huynh2009metrics},
\begin{equation}
    \mathcal{L}_{\text{orientation}} = 1 - \langle q_{\text{predicted}}, q_{\text{ground-truth}} \rangle^2
\end{equation}
for the orientation components in $T^o$ and $T^{p_c}$, where $q$ denotes the 4-dimensional quaternion components of the subgoal transformation/contact pose and $\langle ~, \rangle$ denotes an inner product. We use a binary-cross entropy loss for $X_{mask}$. For the GAT-based models that make $N$ separate predictions of $T^{p_c}$, the reconstruction loss is computed separately for all of the predictions with respect to the single ground truth training example.


\subsection{Evaluation Details}
\label{sec: app-evaluation-details}
\input{figText/gat_pointnet_grasp_feasibility}
\textbf{Global Success Rate Breakdown}
The global success rate used in our single-step skill ablations is designed to aggregate multiple performance factors into a single metric. Here we break these results down to examine the components that lead to the overall trends shown in Figure 4. This section focuses on the \textit{grasp-reorient} skill, as it is most influenced by our particular design choices (similar trends were observed for \textit{push}, while each ablated variation technique worked similarly well for learning the \textit{pull} skill). 

Figures \ref{fig:gat-pointnet-grasp-pos-error} and \ref{fig:gat-pointnet-grasp-ori-error} show the average position and orientation errors over all the single-step trials for variants of the learned samplers. Figures \ref{fig:gat-pointnet-contact-success} and \ref{fig:gat-pointnet-feasibility-success} shows success rate based on whether contact was maintained during skill execution (c) and whether a feasible motion was found within 15 samples from the neural network (d).

From the larger position and orientation errors and reduced sticking success rate, we can see the significant benefit provided by the mask-based subgoal representation (Joint/Indep Mask) over directly using the transformation predicted by the model (Joint Trans). The large position and orientation errors for Joint Trans are typically due to the predicted transformation moving the object to a position vertically above the table. In these cases, the robot moves the object above the table, breaks contact, and the object passively falls onto the table. The large errors in Figure \ref{fig:gat-pointnet-grasp-pos-error} and \ref{fig:gat-pointnet-grasp-ori-error} reflect the  difference between the object's ``in-air'' pose and wherever it ends up on the table after the robot breaks contact. Directly predicting $T^o$ with the Joint Trans architecture also leads to point-cloud configurations that penetrate the table. This penetration frequently causes the robot to lose contact when trying to reach these subgoals because an unexpected contact between the table and the object occurs before skill execution is complete.

The performance between the PointNet++-based and the GAT-based models are supported by the large difference in orientation error, shown in Figure \ref{fig:gat-pointnet-grasp-ori-error}. This is due to the end-effector/object alignment issues depicted in Figure 5a-b, which cause an unexpected object rotation during the initial grasp. We also observe the PointNet++ encoder reduces success rates for sticking and feasibility in Figures \ref{fig:gat-pointnet-contact-success} and \ref{fig:gat-pointnet-feasibility-success}. This lower success rate is because many contact poses predicted by the PointNet++ model lead to a collision with the environment and are more likely to lose contact with the object.

Finally, we see in Figure \ref{fig:gat-pointnet-feasibility-success} that predicting the contact and subgoal parameters jointly (Joint Mask/Trans) instead of independently (Indep Mask) leads to improved Feasibility Success Rate. This is because the Indep models require many more samples to find contact poses and subgoals that work well together. In contrast, the shared representation learned by the Joint models better captures this dependence and helps to more efficiently find a set of parameters that are compatible with each other.

\textbf{Baseline Uniform Sampler Heuristics}
The baselines we compare our learned samplers to in the multi-step evaluation are based on a uniform sampler that uses the point-cloud to sample potential $T^o$ and $T^{p_c}$ values for the \textit{grasp-reorient} and \textit{pull} skills. The samplers are based on the following heuristics: (1) During a \textit{pull}, the palm should contact the object face down at a point that lies on top of the object, (2) During a \textit{grasp-reorient} the palms should face each other and contact antipodal points that are on the side of the object, (3) During a \textit{grasp-reorient} the subgoal $T^o$ should encode a rigid transformation between two stable configurations of the object, that each have different faces contacting the table, (4) the robot should not contact the face of the object that will contact the table in the subgoal configuration.

For the \textit{grasp-reorient} sampler, we first estimate the point-cloud normals and segment all the planes using the Point Cloud Library (PCL)~\cite{Rusu_ICRA2011_PCL}. This process is done once to avoid recomputation in each new sampled configuration; the normals and planes are all transformed together with the overall point-cloud during planning. A plane is sampled and used to solve for $T^o$ using the same registration process described in Section \ref{sec: point-cloud-registration-icp}. Before sampling contact poses, we then filter out points in the point-cloud that are likely to cause infeasibility if contacts were to occur at their location. Specifically, we remove the plane that is used for subgoal registration and the plane opposite to it, since contacting the object on these planes is sure to cause a collision with the table in the subgoal configuration. We also remove points below a $z$ threshold so that the palms are less likely to contact the table in the start configuration. $T^{p_c}$ values are determined from the points that remain after filtering. We sample antipodal points in the remaining object point-cloud for the positional component. For the orientation component, we align the palm-plane normal with the estimated normal at the sampled point and then sample a random angle within the plane by which to rotate the pam. We only consider a range of within-plane palm angles that will not lead the wrist to collide with the table (i.e., the direction of the front of the palm must have a negative $z$ component). 

For the \textit{pull} sampler, we first estimate the point-cloud normals using PCL and then search for a point that has an estimated normal that is close to being aligned with the positive z-axis. This point is used as the position component in $T^{p_c}$. The orientation component is determined by aligning the palm-plane with the positive z-axis in the world frame and then sampling a random angle within the plane by which to rotate the palm.


\subsection{Additional Quantitative Results}
\label{sec: app-additional-results}

\textbf{Multi-step Planning Evaluation with Noisy Point-clouds}
\input{figText/noisy_multistep_table}
To further understand the generalization capabilities of the system, we conducted single-step experiments on 10 novel cylindrical objects with 10 start poses each, using both our learned model and the hand-designed baseline sampler. For the baseline, we approximate the cylinder with an oriented bounding box. We report the global success rate from Figure~\ref{fig: single-step-bar}, and the average position/orientation errors as in Figure~\ref{fig: single-step-breakdown}, with corresponding reference values from the best performing model, tested on cuboids. Figure~\ref{fig: cylinder-single-step-breakdown} shows the results. As expected, we see a drop in performance (larger pose errors and lower success rate) due to the significant distributional shift in the point-cloud observations. However, many of the skill executions are still successful, and performance is still higher than using the baseline sampler with cylinders approximated as cuboids. We hypothesize that the performance gap can be made up by training on a more diverse set of objects. 

\textbf{Single-step Evaluation with Novel Geometric Classes}
\input{figText/cylinder_grasping_quant_results}
To understand how performance differs under more realistic sensor noise, we evaluate our learned models in the multi-step setting described in Section~\ref{sec: multistep-performance-eval} with noisy point-clouds. We use a depth camera noise model taken from \cite{ahn2019depthnoise} to simulate depth image noise, which shows up as noise in the point-cloud that is provided to the samplers. Results shown in Table~\ref{tab: multistep-results-noisy} demonstrate that our system is quite robust to moderate levels of input noise. 

\subsection{System Implementation Details}
\label{sec: app-implementation-details}
\textbf{Model Implementation}
We used PyTorch~\cite{paszke2017automatic} for training and deploying the neural network components of the framework, and we used the PyTorch Geometric library~\cite{torch2019geometric} for implementing the Graph Attention Network and PointNet++ layers in our samplers.

\textbf{Feasibility Checks}
We use the $\texttt{compute\_cartesian\_path}$ capability in the MoveIt!~\cite{moveit} motion planning package for the FeasibleMotion procedure in Algorithm \ref{alg: mutlstep_planning}. This procedure simultaneously converts the cartesian end-effector path computed by the skills into a sequence of joint configurations using RRTConnect \cite{kuffner2000rrt} and checks to ensure the path does not cause singularities or collisions.

\textbf{Contact Pose Refinement}
After $T^{p_c}$ is determined by either the learned or uniform samplers, it is refined based on the dense point-cloud of the object so that the robot is more likely to \textit{exactly} instead of missing contact or causing significant penetration. The refinement is performed by densely searching along the 3D rays aligned with the palm-plane normal, beginning at the position component of $T^{p_c}$, and computing the distance to the closest point in the point-cloud at each point along this line. The point in the point-cloud with the minimum distance is then used to update the positional component of $T^{p_c}$ (the orientation component is unchanged). This process provides a large practical improvement, and we use it for all the skills and all variants of the samplers we evaluated.

\textbf{Real Robot Experiments}
To obtain the real-robot demonstrations of the framework shown in Figure~\ref{fig: real-robot}, we implemented a real-world version of each of the framework's required components. Four Intel RealSense D415 RGB-D cameras were set up in the environment and extrinsically calibrated with respect to the robot's world coordinate system. To obtain the overall desired relative transformation, $T^o_{des}$, we attached an AR tag to the object, moved the object to a start and goal pose in the environment, and calculated the relative transformation between the detected start/goal poses. After obtaining $T^o_{des}$, the AR tag was removed. To segment the object from the overall point-cloud, we first passed the RGB images from each camera to a pre-trained Mask R-CNN~\cite{he2017mask} neural network (using the Detectron2 API~\cite{wu2019detectron2}), applied this mask to the depth-images before obtaining the 3D point-cloud, and finally cropping the point-cloud to a known region near the table to remove outliers in the segmentation mask. After these steps, the planning algorithm was run as described in Appendix~\ref{sec: app-multistep-planning-algo}, and the motions returned by the planner were executed by the robot.

\subsection{Additional Discussion}
\label{sec: app-additional-discussion}

\textbf{Skill Capabilities and Geometry Generalization Implications}
In addition to the explicit assumptions we make in designing our framework, the system also implicitly inherits any additional assumptions made by the underlying skills that are utilized. For the skills from \cite{hogan2020tactile} used in this work, this includes the additional assumption of a limited set of possible types of contact interaction that can occur between the object and the robot/environment. For instance, the contact models used by the skills are not built to plan through rolling, as might occur when pulling a cylindrical object in certain ways. They also do not consider certain 3D interactions, such as toppling, which can happen when contacting a tall thin object during a push skill. 

This interplay of assumptions made by the primitive skills and the higher-level components provided by our framework limited our ability to demonstrate more sophisticated generalization to novel geometries. For instance, multi-step planning and execution on more diverse geometries may require primitive skills that account for more types of contact interactions. Designing such primitives was out of the current scope and is an important direction for future work, but the current framework could be similarly applied if such primitives were available.

\textbf{Point-cloud Occlusion and Observability}
While we used point-cloud obtained from simulated depth cameras in the scene, these point-clouds are relatively complete and unoccluded since we utilize multiple cameras placed at the corners of the table in the scene and do not deal with environments with clutter. Enabling long-horizon manipulation planning in more realistic scenarios where occlusions are significant and less sensors can be used is an important challenge that was beyond the scope of this work. 

Furthermore, the issue of object observability raises the question of whether to deal with planning directly with a heavily occluded observation of the object or to augment the perception system with a shape completion module that predicts the geometry in the occluded regions, as is done in other manipulation systems that deal with limited observability \cite{van2020learning}. Directly working with the partial point-cloud would present numerous challenges for our system (and any manipulation planning framework). This is because our manipulation action parameters are represented explicitly with respect to points on the point-cloud since these denote locations where contact on the object can confidently be made with the robot or with the environment. Therefore, we hypothesize shape completion techniques to be more suitable and of great importance for extending the ideas presented in this work to scenarios where objects are more occluded than in this paper.

\textbf{Failure Modes}
\input{figText/baseline_compare/failure_table}
\input{figText/baseline_compare/samples_table}
To understand the performance gap between our learned samplers and the baselines, we tracked the frequency of different types of failures during planning (Table~\ref{tab: baseline-compare-failure}) and the average number of sampling iterations performed by both methods (Table~\ref{tab: baseline-compare-samples}). We found that the main issue with the baseline is that it spends more time checking the feasibility of infeasible motions due to sampling many contact/subgoals that are infeasible either for the robot, or with each other, or both. In this way, we can interpret part of the value provided by our samplers as having learned the helpful biases that guide sampling toward regions that are likely to be feasible, similar to as in other works that learn biased samplers for motion planning and TAMP \cite{chitnis2016guided, beomjoon2018sampler}. We found that computation efficiency between both methods is similar (see Table~\ref{tab: baseline-compare-samples}).

Failure modes that lead to poor execution of the skills after parameters are sampled or multi-step plans are found are primarily related to small errors in predicted palm poses or subgoal transforms that can lead to unintended outcomes. This includes predicting contact that is too soft so that the object slips or predicting contact that penetrates the object too much, which causes large internal forces and robot joint torques along with highly unrealistic physical behavior in the simulator. Additionally, as is usual with multi-step scenarios, errors can accumulate between iterative predictions during planning. A common example is a small error in a subgoal prediction that leads the transformed point-cloud to be in a slightly unstable configuration that is not reflective of the configuration the object settles into when the skill is executed. This instability can lead the following skill to be executed poorly due to the nominal plan expecting the object to be in a different configuration than what it actually reached.

Finally, many failure modes we observe are generally related to the realities of running the obtained plans and low-level motions purely open-loop. Ideally, high-level replanning/plan refinement based on updated point-cloud locations at each step can be used to mitigate cascading error issues described above, and controllers developed in conjunction with the primitive skill planners \cite{hogan2020tactile} could be used to make local adjustments based on feedback from the contact interface to better enforce the sticking contact assumption to be true when a skill is executed.

\textbf{Real-world/Real-robot Implementation Considerations}
Implementing the primitive skills and the framework proposed in this work on a real robotic platform requires certain considerations. Practical difficulties can arise when implementing the plans on a stiff position-controlled platform without any sensing to guide the skill execution. Stiff position control creates difficulties because skills like pulling and grasping are fundamentally designed to apply compressive forces on the object during execution, which can be problematic when dealing with near-rigid object and large position gains. 

We found a large practical benefit introducing a passively compliant element at the wrist of the robot that deforms to take up forces encountered when commanding contact pose positions that slightly penetrate the object (this compliance was also modeled in the PyBullet YuMi simulation). Other ways of introducing non-stiff behavior can be similarly applied to realize the behaviors that are planned by the manipulation skills, such as hybrid position/force control or cartesian impedance control. Small errors in commanded positions can also lead the skill to miss making contact entirely. Guarded movements using aggregated wrist force/torque sensing or local contact sensing can help alleviate execution errors of this type (i.e., by moving to a nominal pose predicted by the sampler and slowly approaching the object in the direction of the palm normal until a certain force threshold is reached).


%% file: algText/planning_alg.tex
\begin{algorithm}
\caption{Multistep Planning}
\label{alg: mutlstep_planning}
  \begin{algorithmic}[1]
\State \textbf{Input:}  Desired transformation $T^o_{des}$, Point-cloud $X$, Plan skeleton $PS$, Number of skeleton steps $T$, Buffers for each step in skeleton $\{\mathcal{B}_t\}_{t=1}^{T}$, Skill parameter sampling distribution for the skill at each step in skeleton $\{p_{\pi_t}(\cdot | X)\}_{t=1}^{T}$, Maximum number of samples at each step in the skeleton $K_{max}$ \hspace{20mm}
\vspace{10pt}
\State root\_node $\leftarrow$ InitPointCloudNode($X, I^4$, None, None) \Comment{Initialize root node with initial point-cloud, identity transformation, no contact pose, and no parent}
\State $\mathcal{B}_{1},...,\mathcal{B}_{T} \leftarrow \emptyset$ \Comment{Empty skill buffers}
\State $\mathcal{B}_{1} \leftarrow \mathcal{B}_{1} ~\cup$ root\_node
\State done $\leftarrow$ False
\While{not done}
    \For{skill step $t$ in 1,...,$T$}
    \State $k \leftarrow 1$ \Comment{If feasible parameters not found after $K_{max}$ samples, move to next step}
    \While{$k < K_{max}$} 
        \State node $ \sim \mathcal{B}_t(\cdot)$ \Comment{Sample start state from corresponding buffer}
        \State $X \leftarrow$ node.pointcloud
            \If{PreconditionsSatisfied($X$)} \Comment{Check if point-cloud is valid for sampling}
                \State $T^o, T^{p_c} \sim p_{\pi_t}(\cdot | X)$ \Comment{Use point-cloud to sample from skill model}
                \If{$t$ equals $T$}
                    \State $T^o \leftarrow$ GetFinalTransformation(node)
                    \If{FeasibleMotion($\pi_T, T^o, T^{p_c}$)}
                        \State $X_{final} \leftarrow$ TransformPointCloud($T^o, X$)
                        \State final\_node $\leftarrow$ InitPointCloudNode($X_{final}, T^o, T^{p_c}$, node)
                        \State done $\leftarrow$ True
                    \EndIf
                \Else
                    \If{FeasibleMotion($\pi_t$, $T^o, T^{p_c}$)}
                        \State $X_{new} \leftarrow$ TransformPointCloud($T^o, X$)
                        \State new\_node $\leftarrow$ InitPointCloudNode($X_{new}, T^o, T^{p_c}$, node)
                        \State $\mathcal{B}_{t+1} \leftarrow \mathcal{B}_{t+1} \cup$ new\_node \Comment{Add to buffer for next step}
                    \EndIf
                \EndIf
            \EndIf
        \State $k \leftarrow k + 1$  
    \EndWhile
    \EndFor
    \If{timed out}
        \State \Return None
    \EndIf
\EndWhile
\State plan $\leftarrow$ ReturnPlan(final\_node) \Comment{Backtrack through parents from final node}
\State \Return{plan $(T^{p_c}_{1:T}, T^o_{1:T})$}
\end{algorithmic}
\end{algorithm}

%% file: figText/gat_architectures.tex
\begin{figure}
    \centering
  \begin{tabular}{ccc}
     \begin{subfigure}[b]{0.4\linewidth}
		\includegraphics[width=1.0\linewidth]{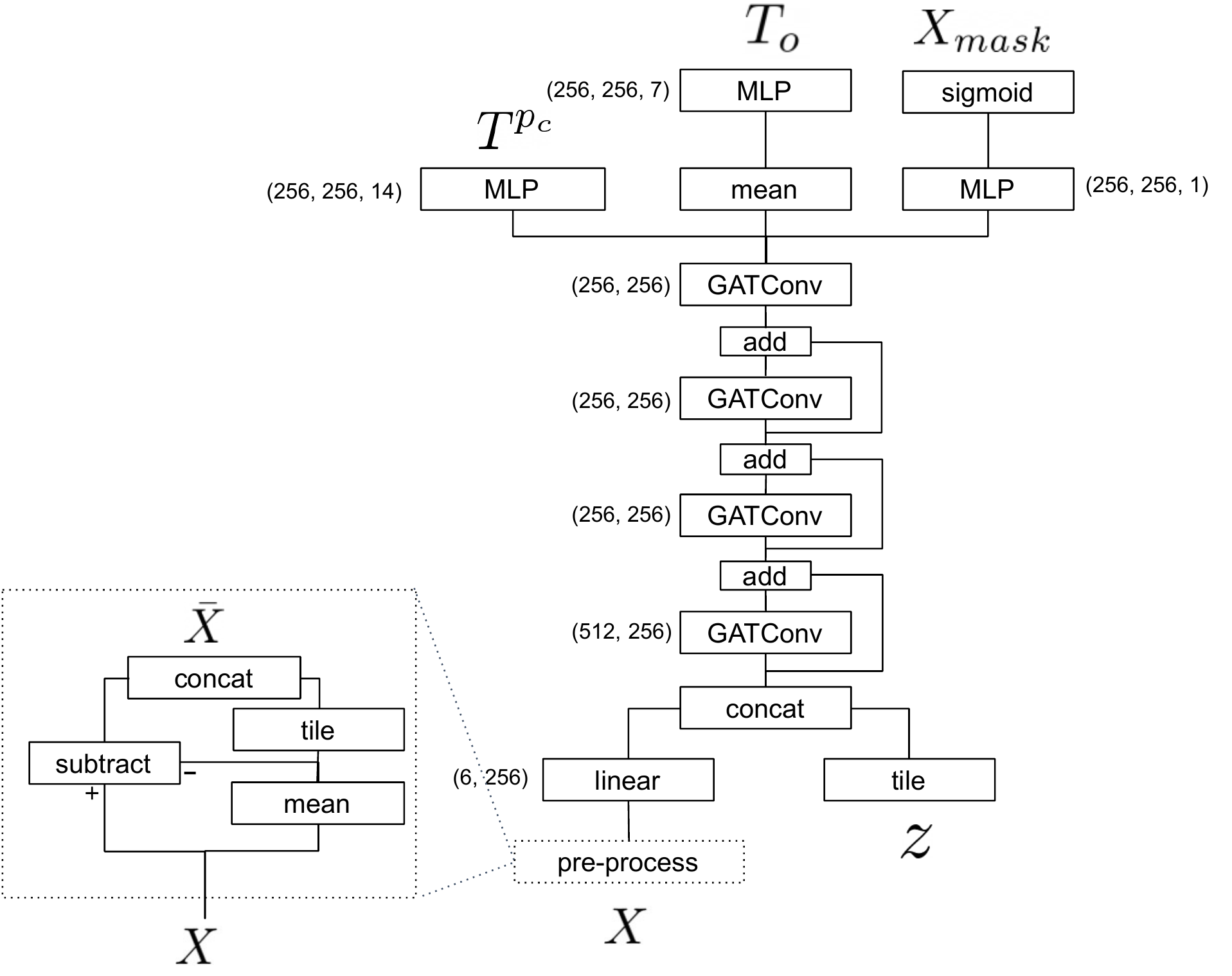}
		\caption{Joint decoder} 
		\label{fig:gat-joint-decoder}
	\end{subfigure}\hfill&
     \begin{subfigure}[b]{0.245\linewidth}
		\includegraphics[width=1.0\linewidth]{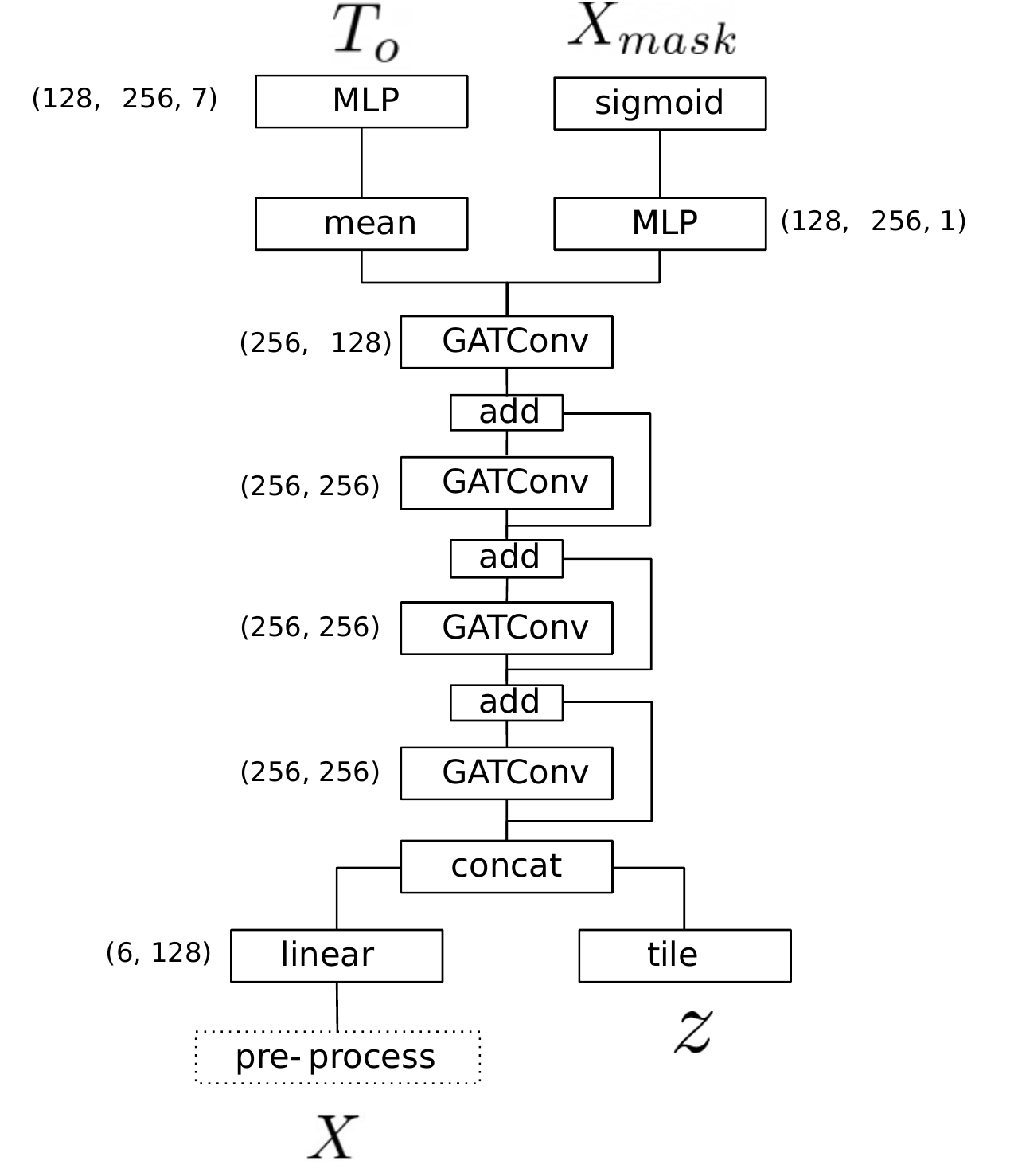}
		\caption{Subgoal decoder} 
		\label{fig:gat-indep-subgoal-decoder}
	\end{subfigure}\hfill&
     \begin{subfigure}[b]{0.21\linewidth}
		\includegraphics[width=1.0\linewidth]{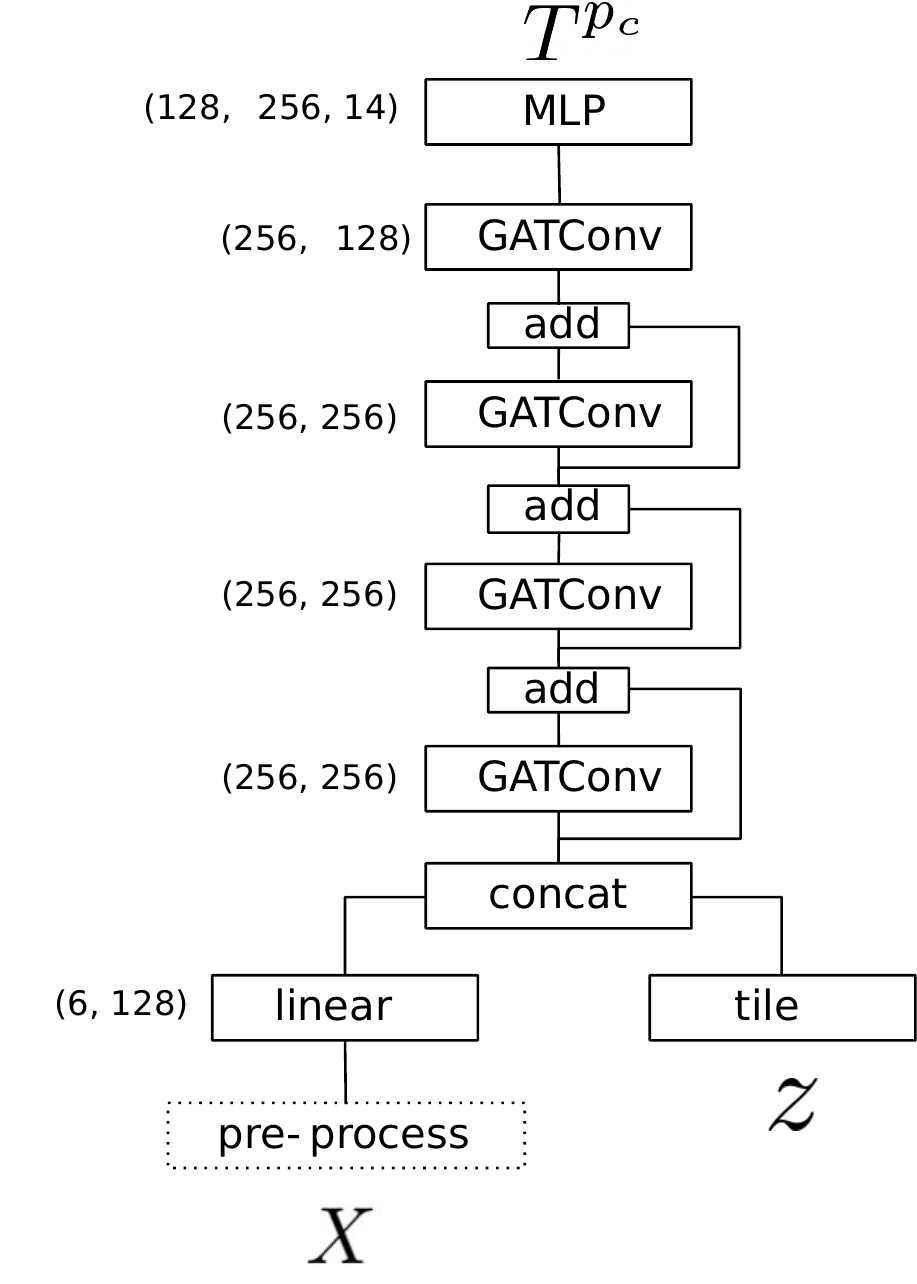}
		\caption{Contact decoder} 
		\label{fig:gat-indep-contact-decoder}
	\end{subfigure}
	\vspace{10pt}
\\
    \begin{subfigure}[b]{0.35\linewidth}
		\includegraphics[width=1.0\linewidth]{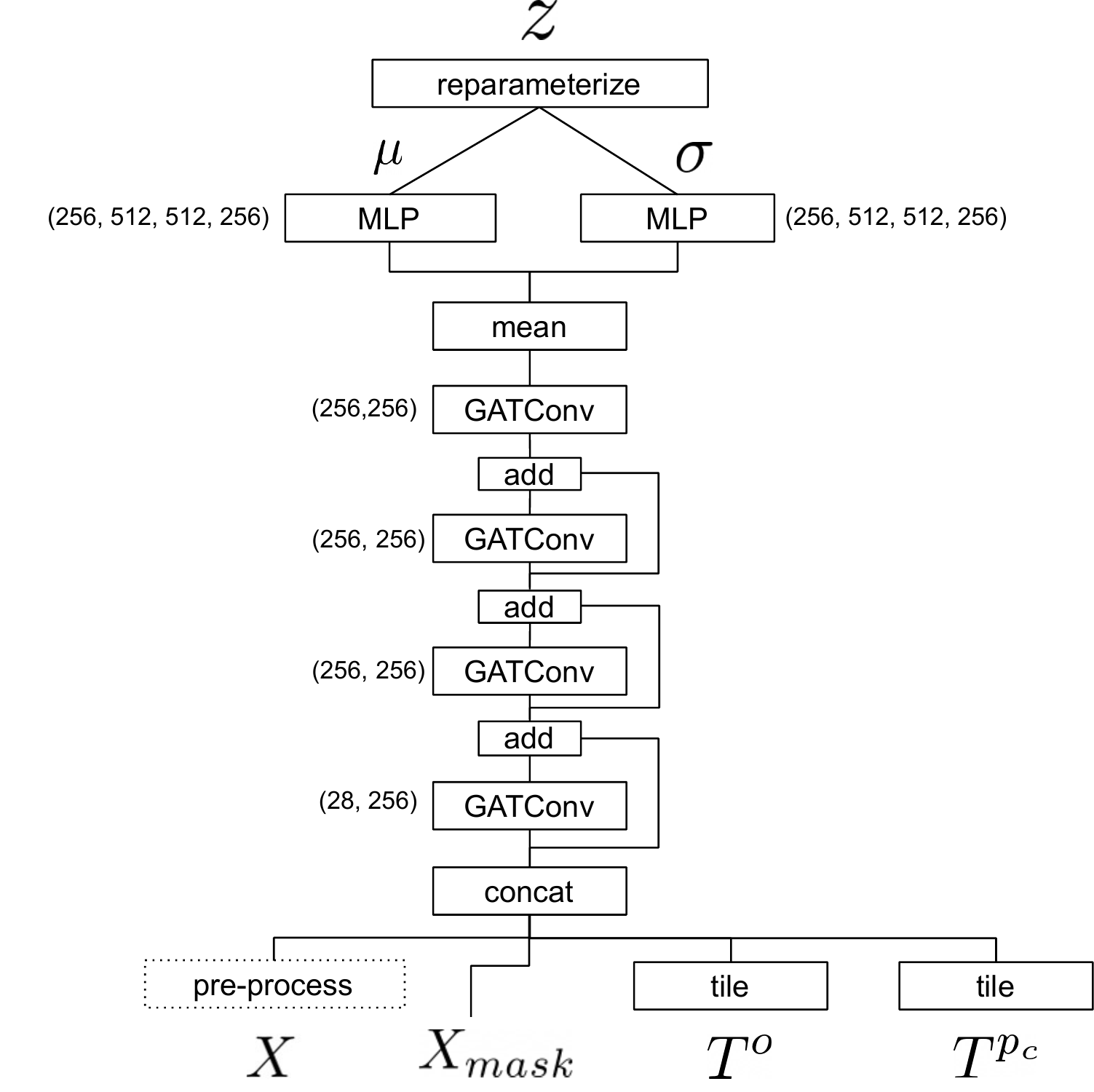}
		\caption{Joint encoder}
		\label{fig:gat-joint-encoder}
    \end{subfigure}\hfill&
    \begin{subfigure}[b]{0.3\linewidth}
		\includegraphics[width=1.0\linewidth]{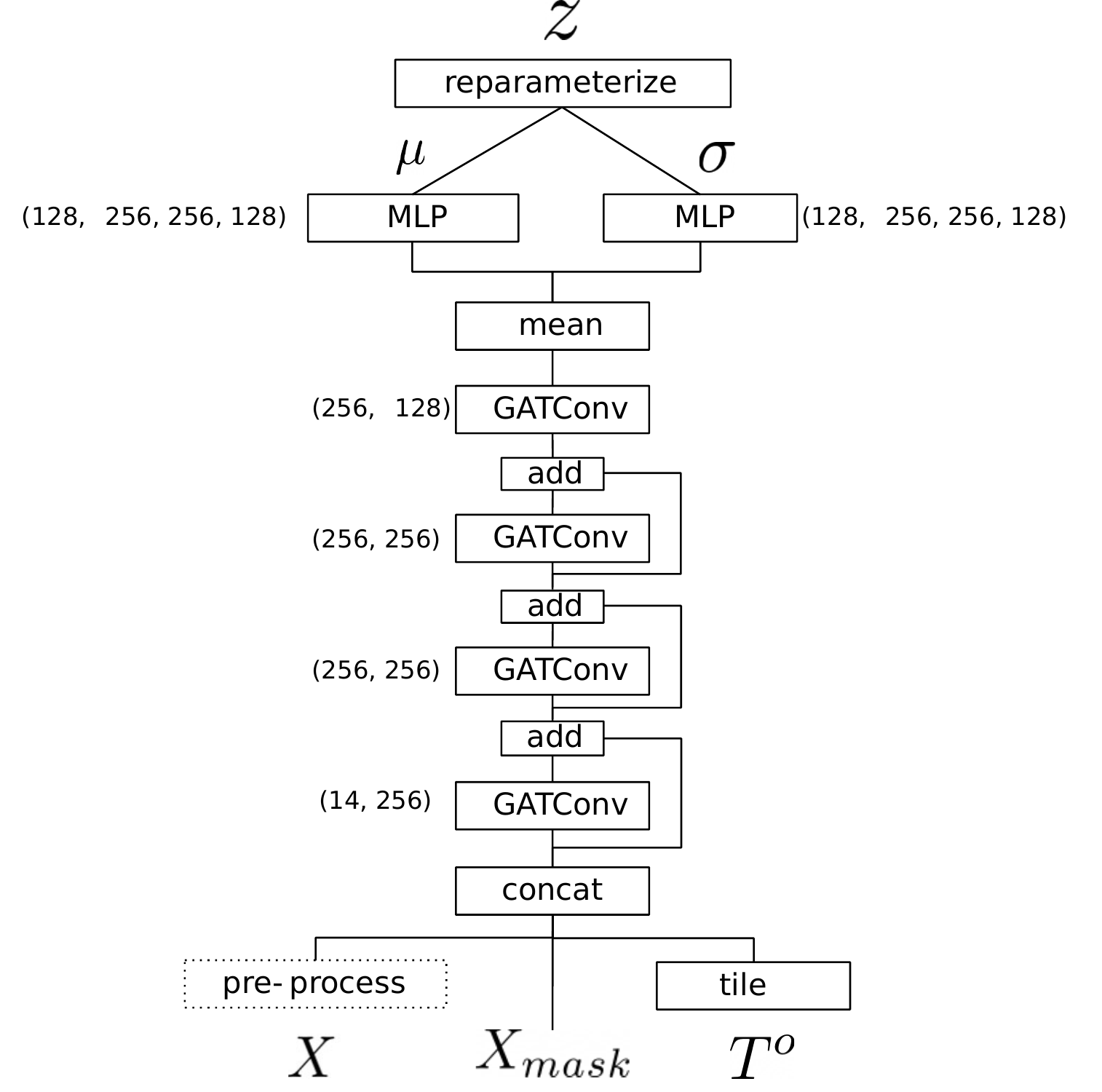}
		\caption{Subgoal encoder}
		\label{fig:gat-indep-subgoal-encoder}
    \end{subfigure}\hfill&
    \begin{subfigure}[b]{0.3\linewidth}
		\includegraphics[width=1.0\linewidth]{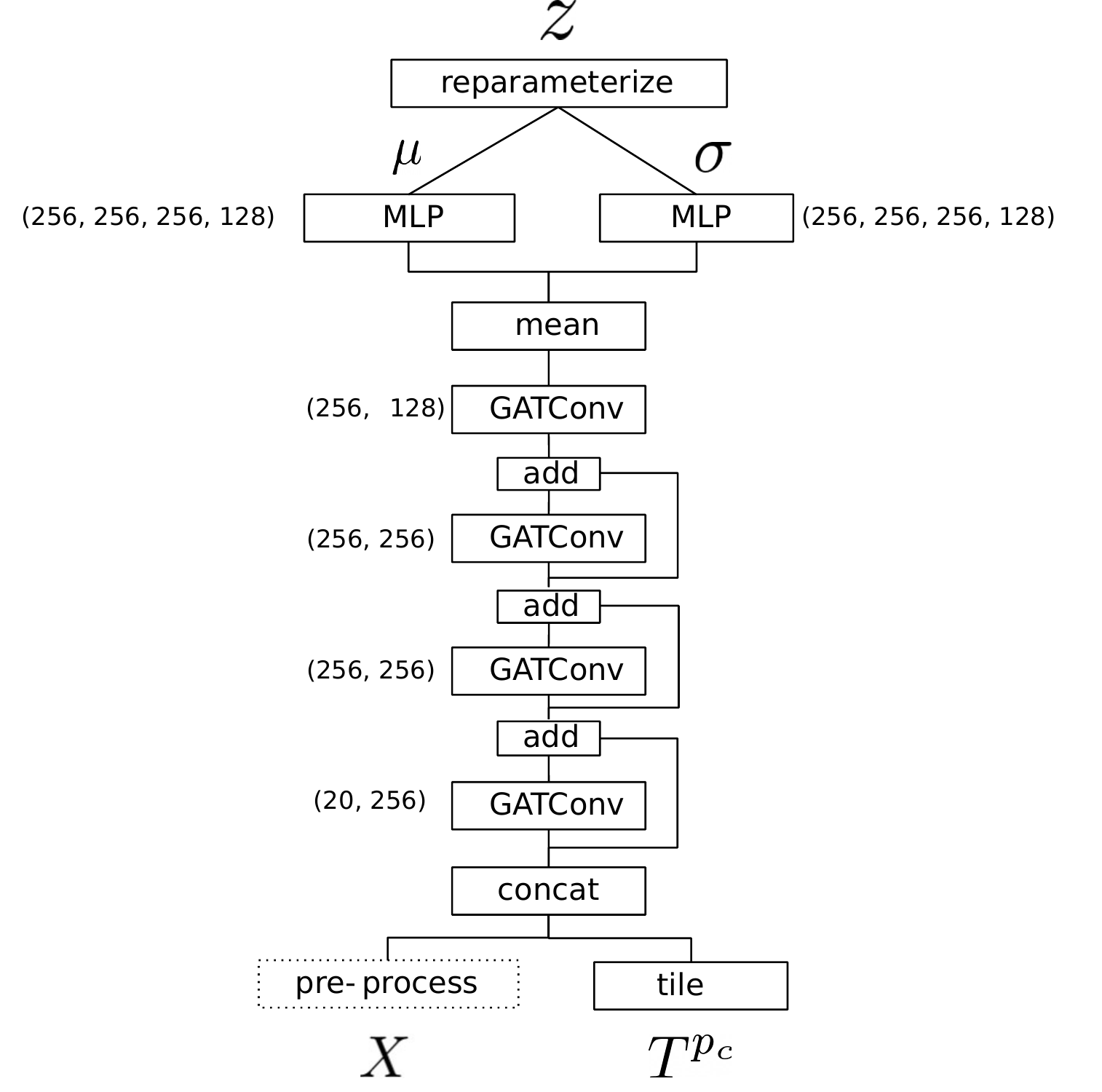}
		\caption{Contact encoder}
		\label{fig:gat-indep-contact-encoder}
    \end{subfigure}
    \end{tabular}    
    \caption{Network architectures based on Graph Attention Networks \cite{velivckovic2017graph} for joint (a,d) and independent (b-c,e-f) CVAE decoder and encoder. The pre-processing block for all architectures is identical to the one shown in (a). GATConv layers perform self-attention over nodes in a neighborhood to propagate node features (see \cite{velivckovic2017graph} for details). We represent the input point-cloud as a fully connected graph.}
    \label{fig: gat-architecture-detailed}
    \vspace{-10pt}
\end{figure}

%% file: figText/pointnet_architectures.tex
\begin{figure}
    \centering
  \begin{tabular}{ccc}
     \begin{subfigure}[b]{0.35\linewidth}
		\includegraphics[width=1.0\linewidth]{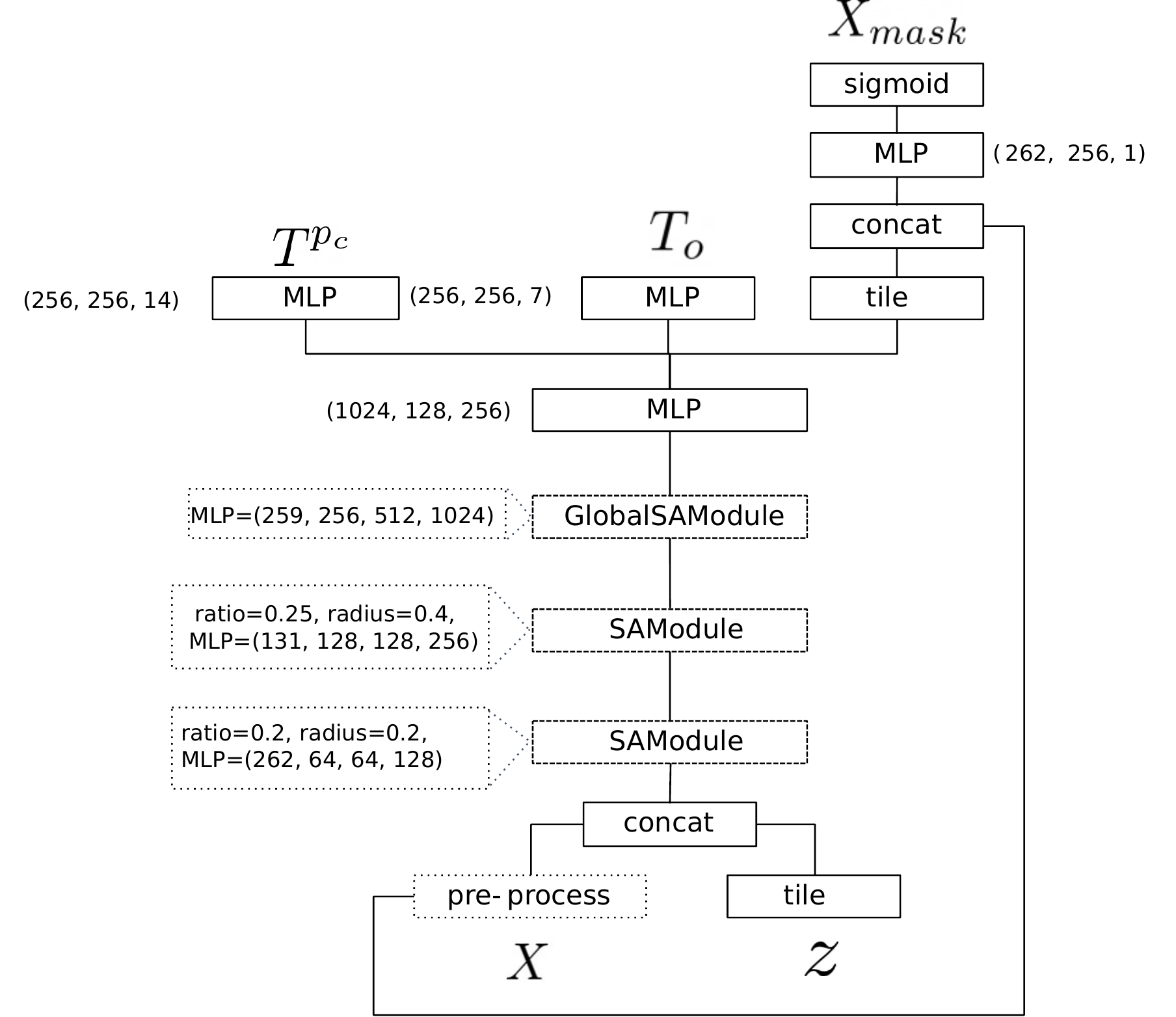}
		\caption{Joint decoder} 
		\label{fig:pn-joint-decoder}
	\end{subfigure}&
     \begin{subfigure}[b]{0.25\linewidth}
		\includegraphics[width=1.0\linewidth]{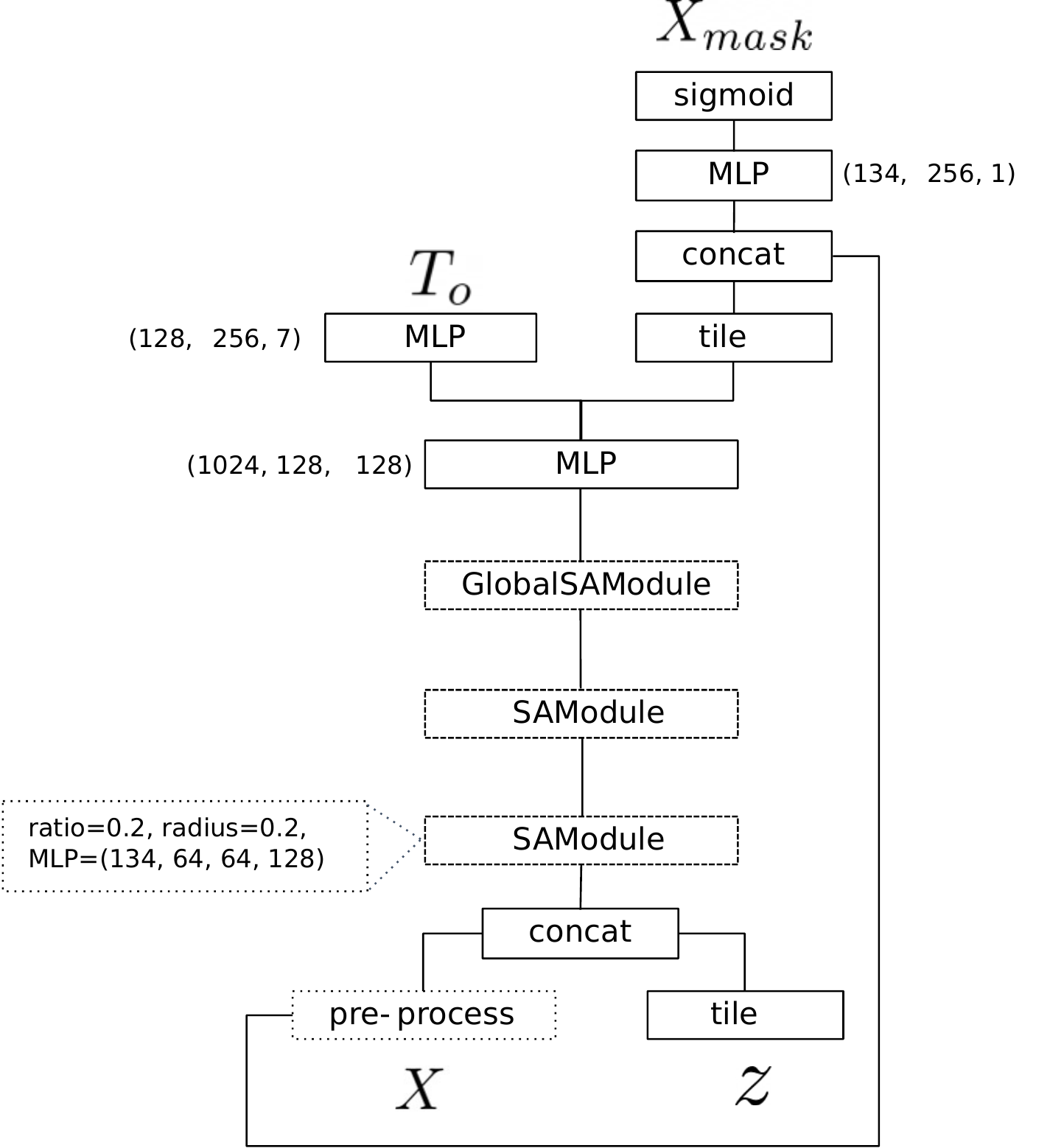}
		\caption{Subgoal decoder} 
		\label{fig:pn-indep-subgoal-decoder}
	\end{subfigure}&
     \begin{subfigure}[b]{0.25\linewidth}
		\includegraphics[width=1.0\linewidth]{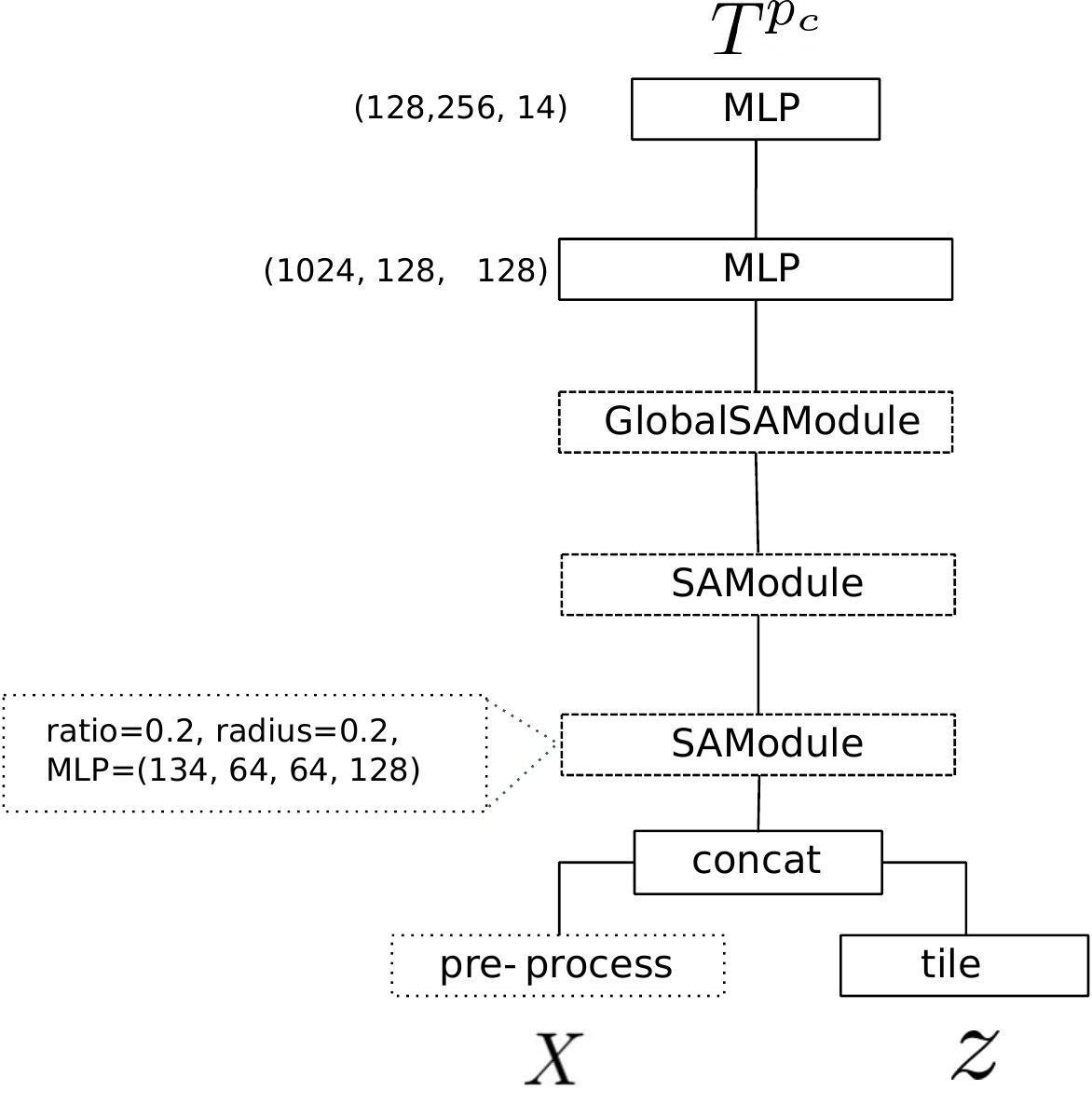}
		\caption{Contact decoder} 
		\label{fig:pn-indep-contact-decoder}
	\end{subfigure}\\
    \begin{subfigure}[b]{0.32\linewidth}
		\includegraphics[width=1.0\linewidth]{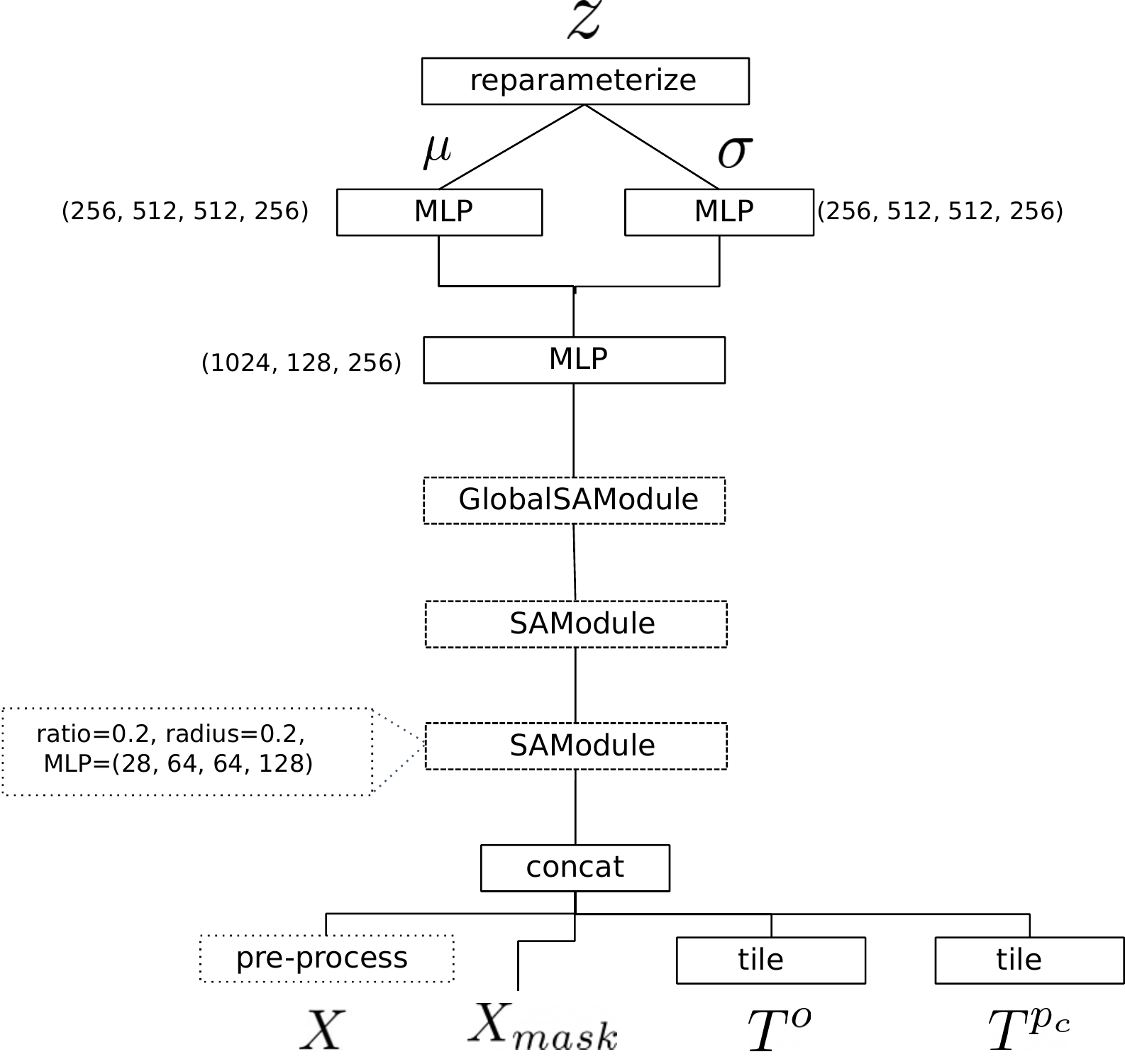}
		\caption{Joint encoder}
		\label{fig:pn-joint-encoder}
    \end{subfigure}&
    \begin{subfigure}[b]{0.29\linewidth}
		\includegraphics[width=1.0\linewidth]{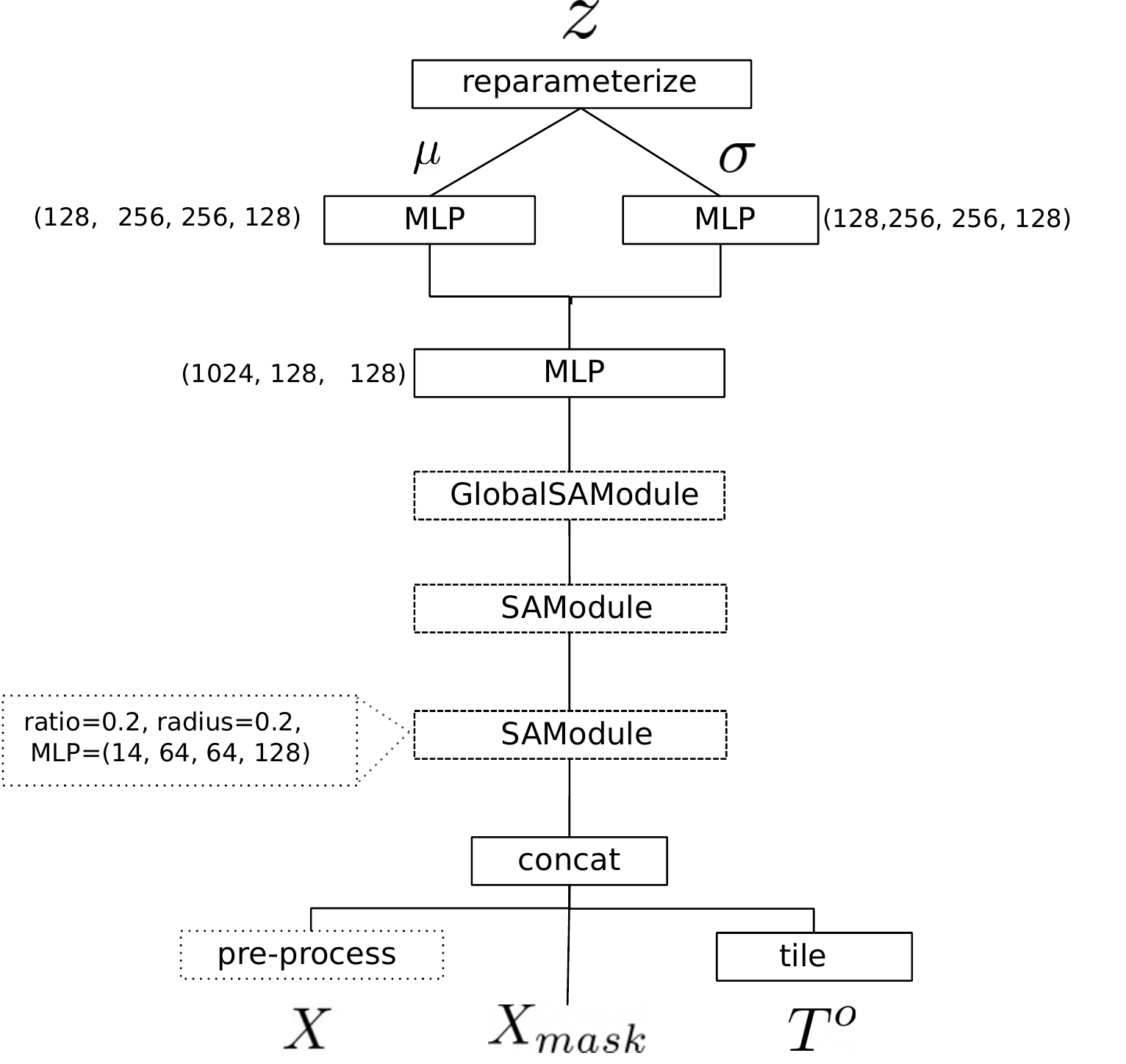}
		\caption{Subgoal encoder}
		\label{fig:pn-indep-subgoal-encoder}
    \end{subfigure}&
    \begin{subfigure}[b]{0.32\linewidth}
		\includegraphics[width=1.0\linewidth]{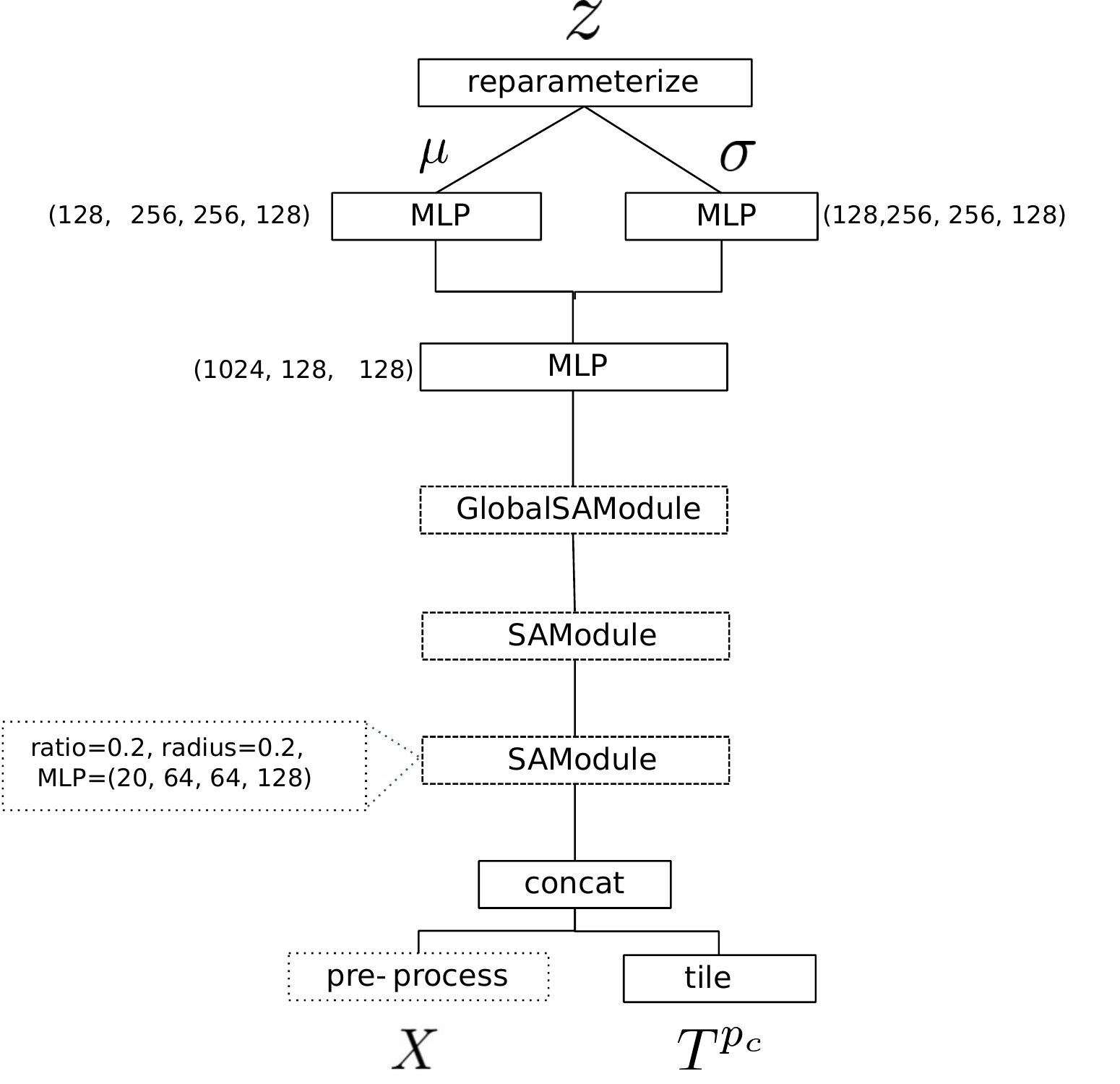}
		\caption{Contact encoder}
		\label{fig:pn-indep-contact-encoder}
    \end{subfigure}
    \end{tabular}    
    \caption{Network architectures based on PointNet++ \cite{qi2017pointnet++} for joint (a,d) and independent (b-c,e-f) CVAE decoder and encoder. SAModules perform the hierarchical set abstraction operation in the PointNet++ (for details see \cite{qi2017pointnet++}). The only difference between the initial SAModules in the various models is the input dimension.}
    \label{fig: pointnet-architecture-detailed}
    \vspace{-10pt}
\end{figure}

%% file: figText/training_table.tex
\def\arraystretch{1.0}
\begin{table}[t]
\centering
\caption{Training data statistics} 
\begin{tabular}{c|cc}
\hline \hline
Skill Type & Number of objects & Number of samples \\ \hline 
\textit{Pull} & 110 & 12712 \\
\textit{Grasp-Reorient} & 64 & 8408 \\
\textit{Push} & 223 & 3835 \\
\hline \hline
\end{tabular}
\label{tab: training-data-statistics}
\end{table}

%% file: figText/gat_pointnet_grasp_feasibility.tex
\begin{figure}
    \centering
  \begin{tabular}{ccc}
    \begin{subfigure}{0.23\linewidth}
		\includegraphics[width=1.0\linewidth]{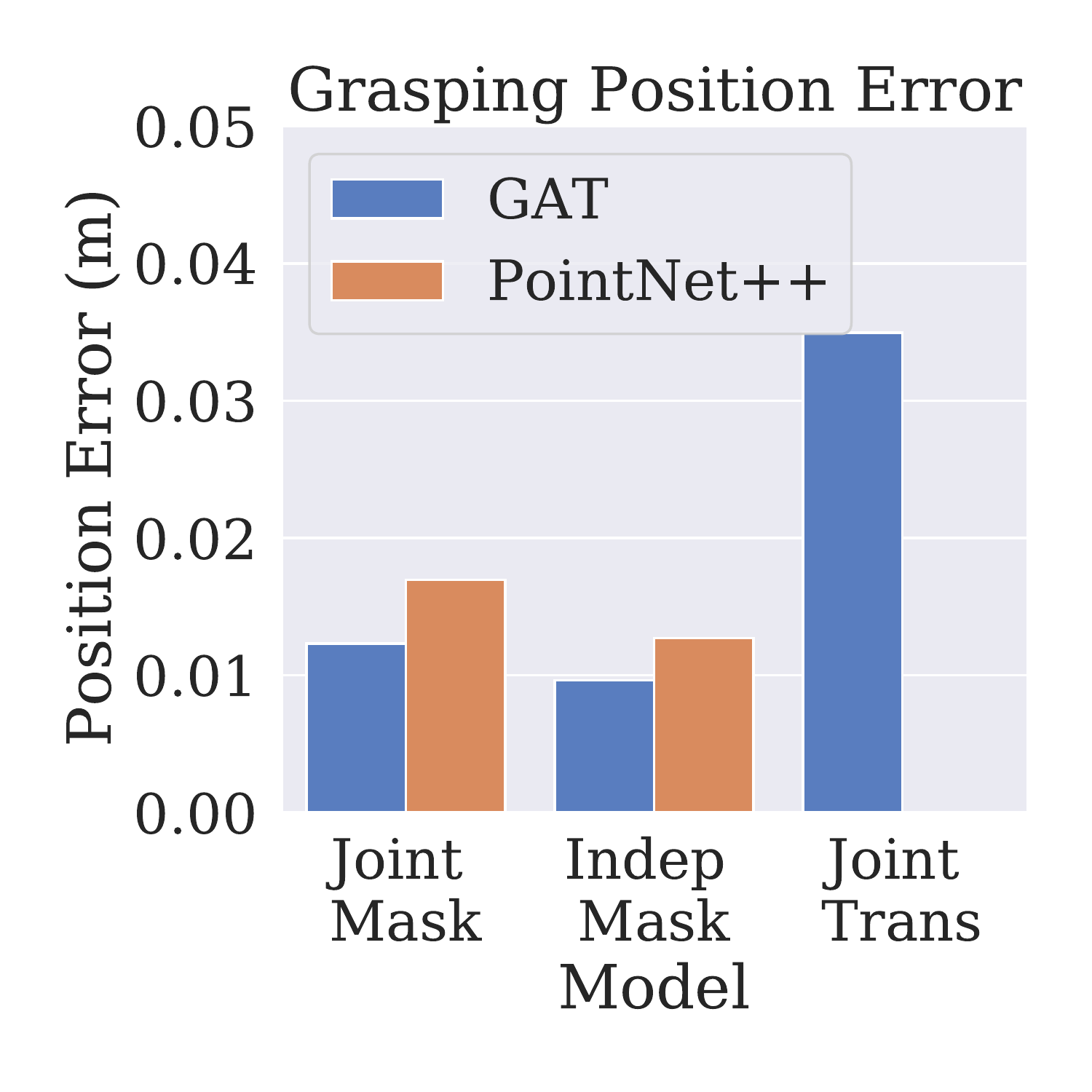}
		\caption{} 
		\label{fig:gat-pointnet-grasp-pos-error}
	\end{subfigure}\hfill&
    \begin{subfigure}{0.23\linewidth}
		\includegraphics[width=1.0\linewidth]{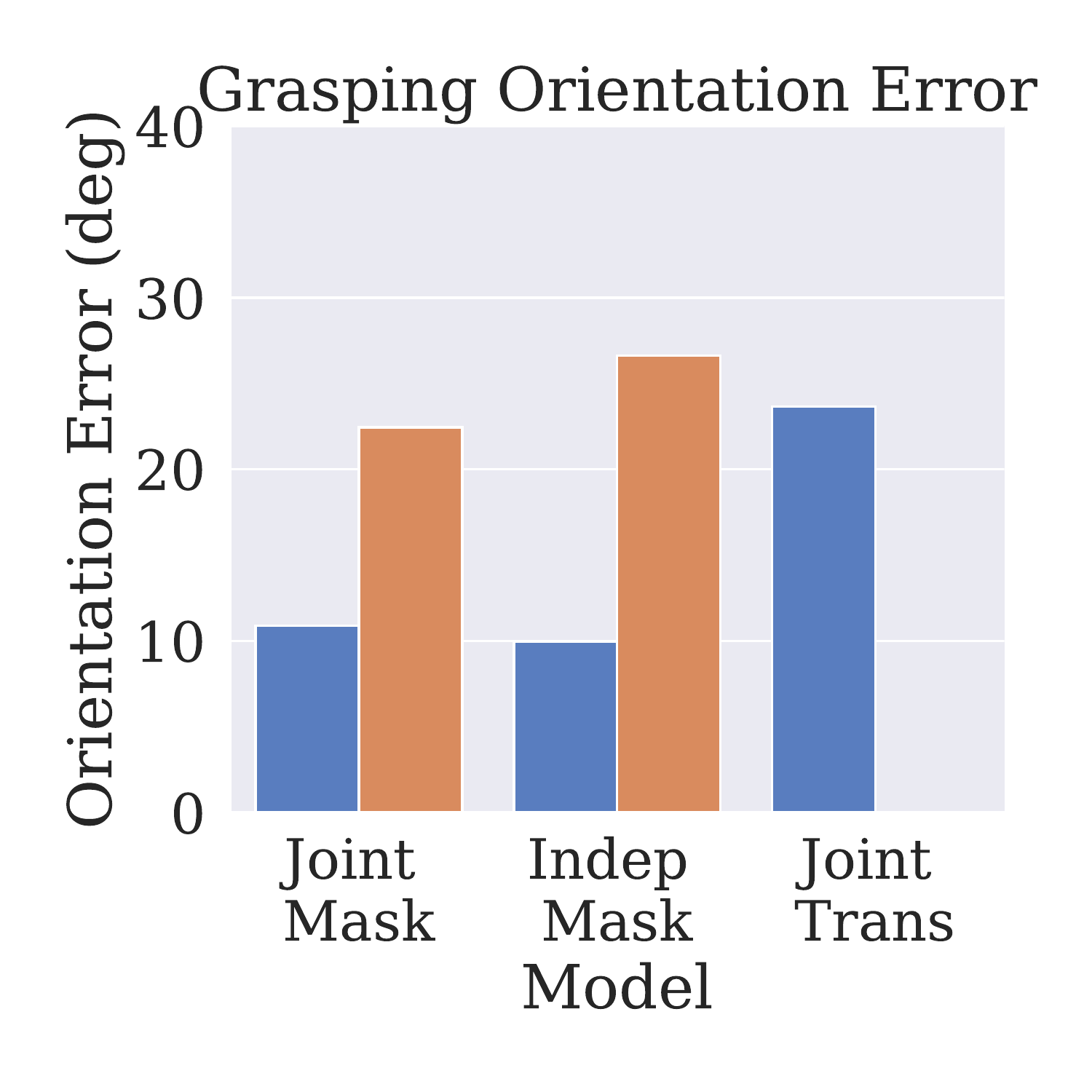}
		\caption{}
		\label{fig:gat-pointnet-grasp-ori-error}
    \end{subfigure}\hfill&
    \begin{subfigure}{0.23\linewidth}
		\includegraphics[width=1.0\linewidth]{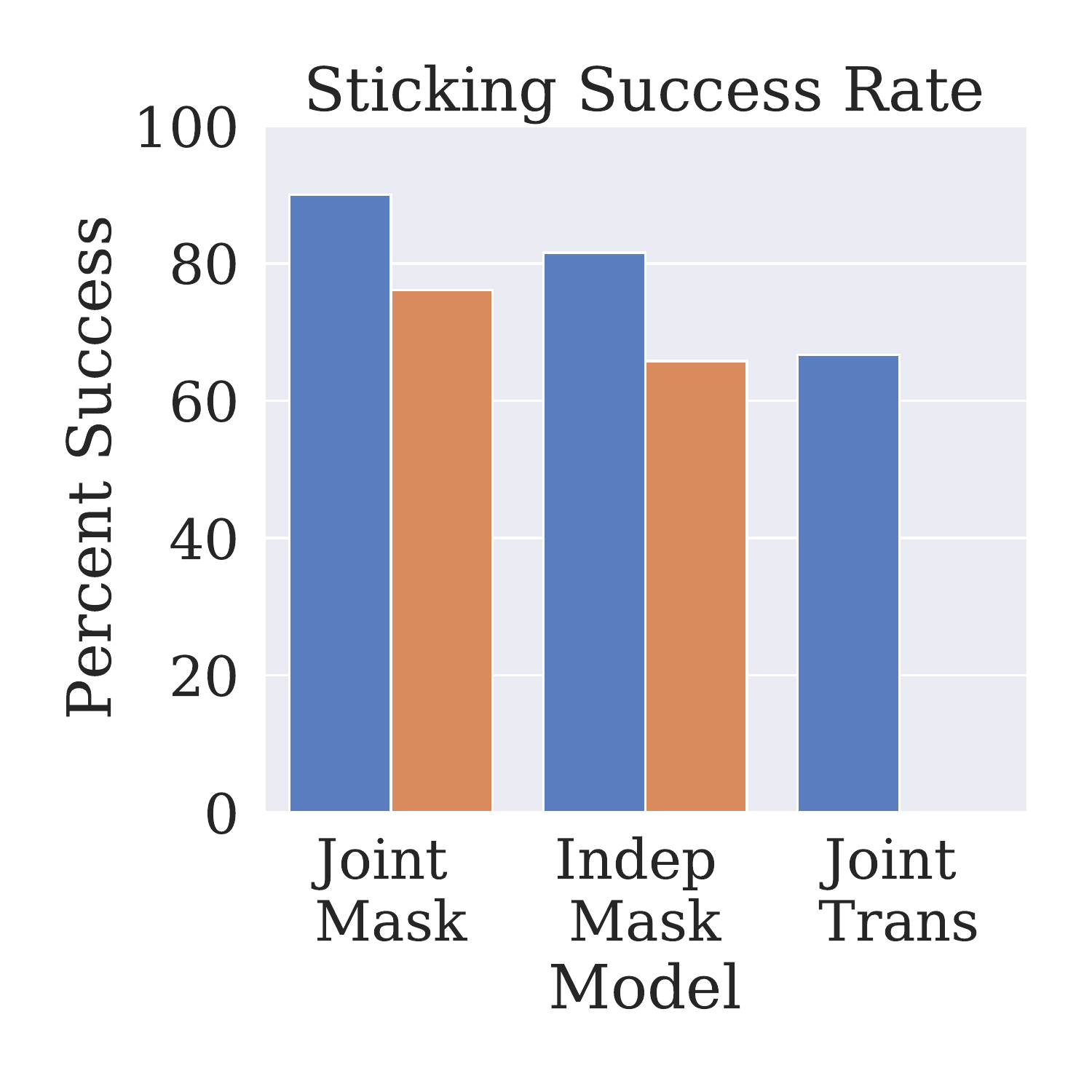}
		\caption{} 
		\label{fig:gat-pointnet-contact-success}
	\end{subfigure}\hfill
    \begin{subfigure}{0.23\linewidth}
		\includegraphics[width=1.0\linewidth]{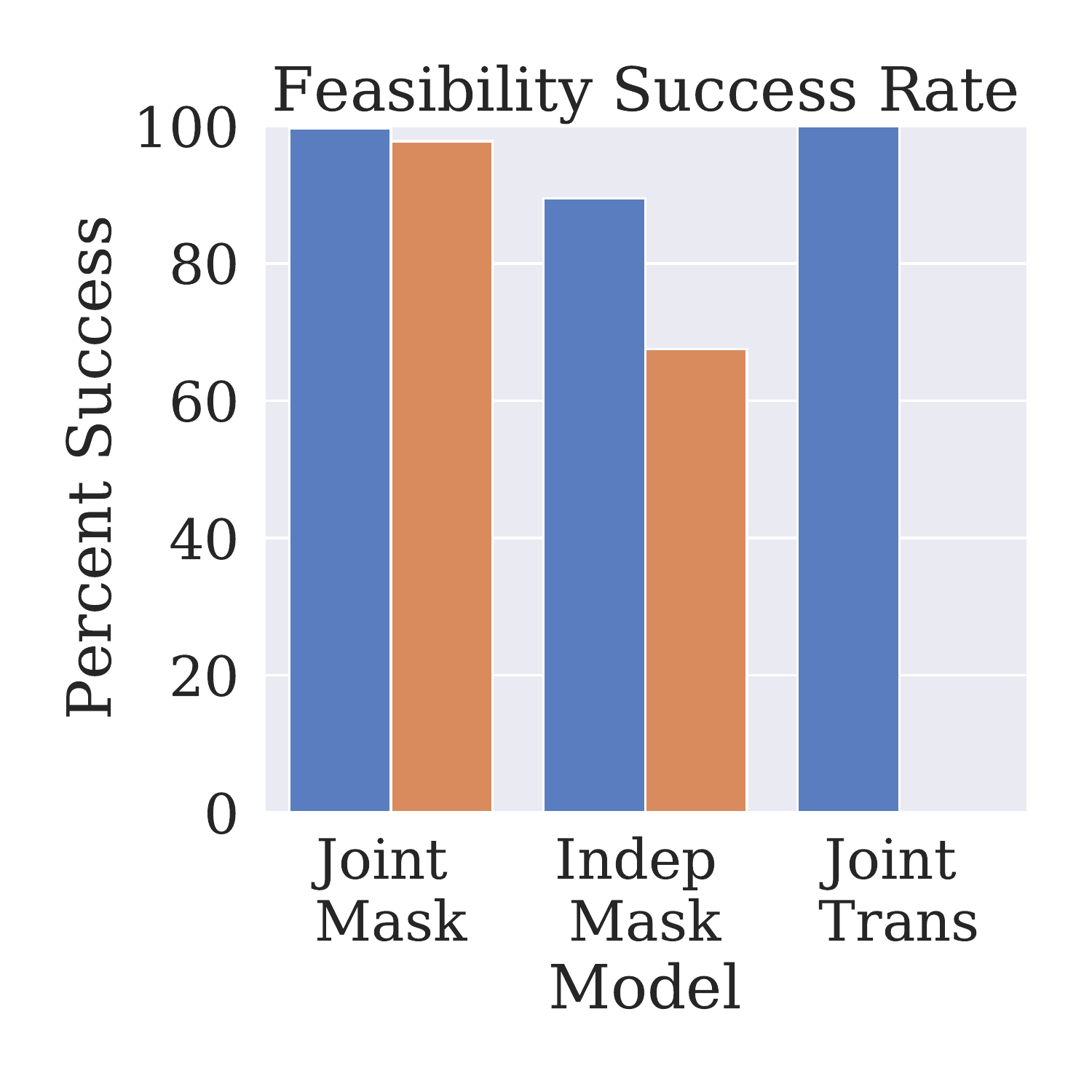}
		\caption{} 
		\label{fig:gat-pointnet-feasibility-success}
	\end{subfigure}\hfill

    \end{tabular}    
    \caption{More detailed single-step ablation results for \textit{grasp-reorient} skill. (a) and (b) show the average position (a) and orientation (b) error between the subgoal transformation provided to the skill and the actual transformation the object underwent between its initial configuration and final configuration after the skill was executed. (c) shows the fraction of trials where a contact was maintained with the object through the duration of the reorientation and (d) shows the fraction of trials where a feasible motion was found within 15 samples from the learned model.}
    \label{fig: single-step-breakdown}
    \vspace{-10pt}
\end{figure}

%% file: figText/noisy_multistep_table.tex
\def\arraystretch{1.0}
\begin{table}[t]
\small
\centering
\caption{Multistep planning results on noise-corrupted point-clouds using learned samplers trained on point-clouds without noise, similar to Table~1. The depth camera noise model was obtained from \cite{ahn2019depthnoise}} 
\begin{tabular}{cccc}
\hline \hline
Skeleton & Success Rate (\%) & Pose Error (cm / deg) & Time (s) \\ \hline 
\texttt{PG} & 94.5 $\pm$ 3.3 & 1.1 $\pm$ 0.5 ~/~ 5.8 $\pm$ 3.0 & 30.0 $\pm$ 5.3 \\
\texttt{GP} & 95.9 $\pm$ 2.8 & 3.9 $\pm$ 0.5 ~/~ 30.1 $\pm$ 5.7 & 22.8 $\pm$ 5.3 \\
\texttt{PGP} & 87.9 $\pm$ 4.5 & 2.1 $\pm$ 0.3 ~/~ 16.6 $\pm$ 3.7 & 55.5 $\pm$ 10.4\\
\hline \hline
\end{tabular}
\label{tab: multistep-results-noisy}
\end{table}

%% file: figText/cylinder_grasping_quant_results.tex
\begin{figure}
    \centering
  \begin{tabular}{ccc}
    \begin{subfigure}{0.315\linewidth}
		\includegraphics[width=1.0\linewidth]{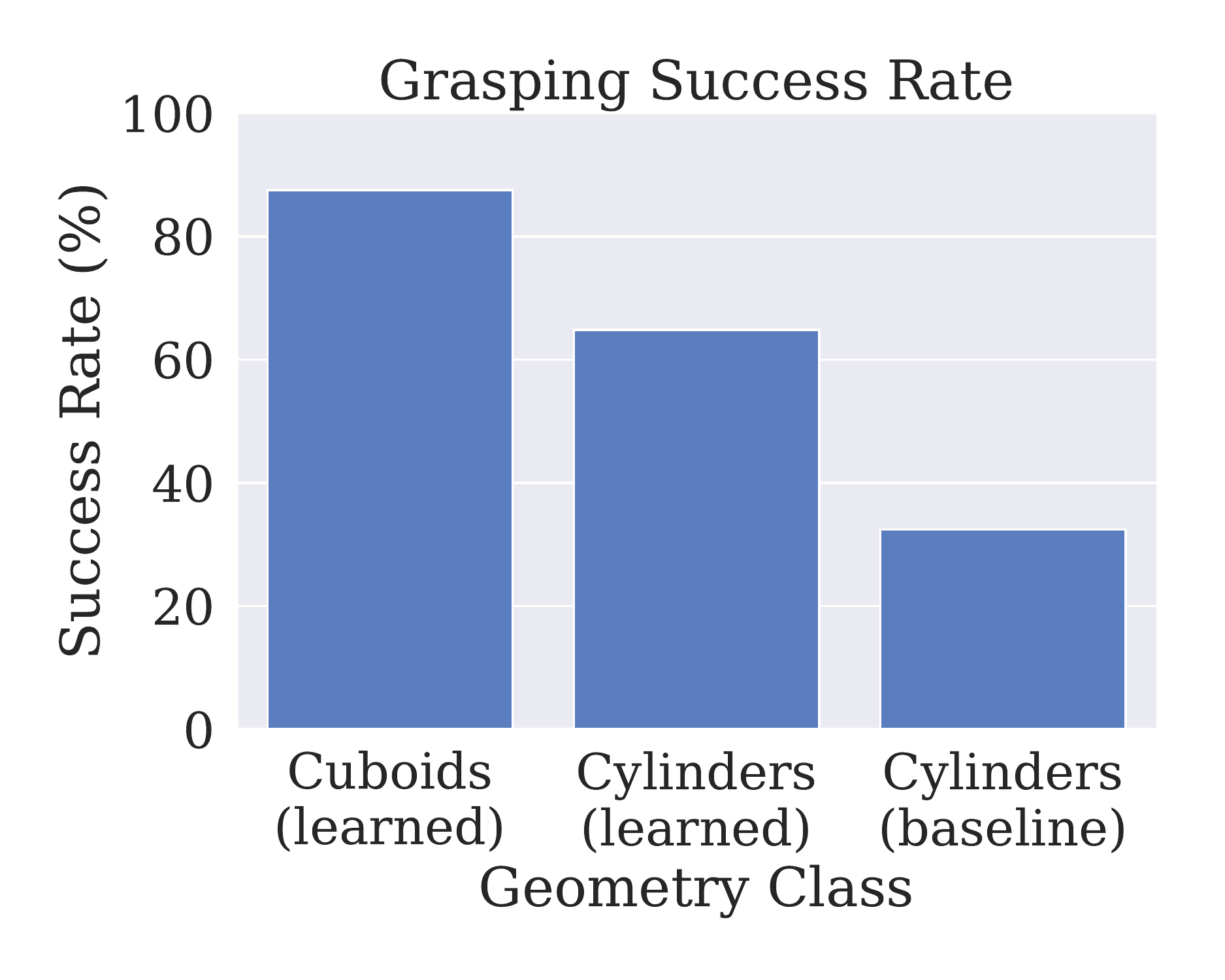}
		\caption{} 
		\label{fig:cuboid-cylinder-success-rate}
	\end{subfigure}\hfill&
    \begin{subfigure}{0.315\linewidth}
		\includegraphics[width=1.0\linewidth]{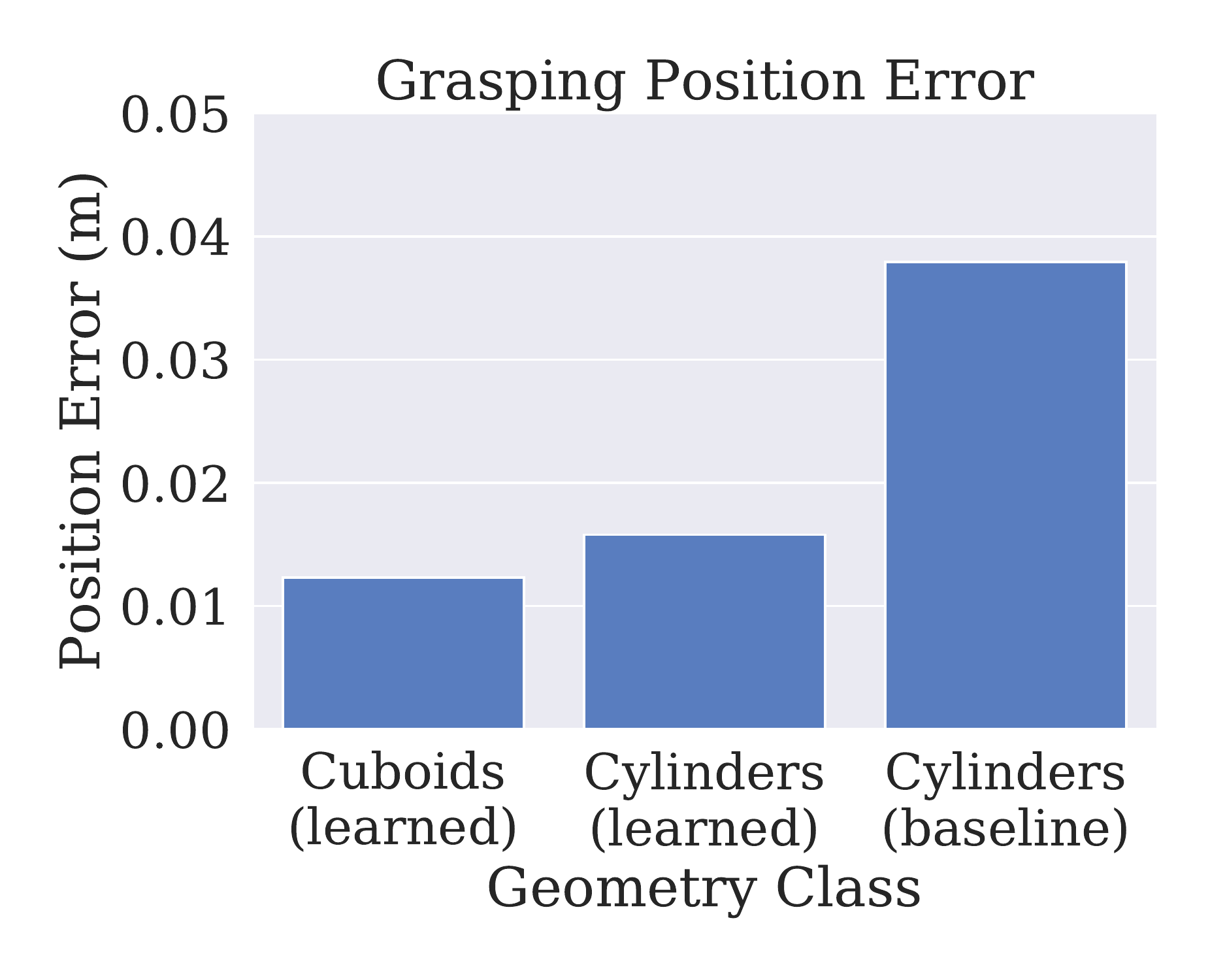}
		\caption{}
		\label{fig:cuboid-cylinder-pos-error}
    \end{subfigure}\hfill&
    \begin{subfigure}{0.315\linewidth}
		\includegraphics[width=1.0\linewidth]{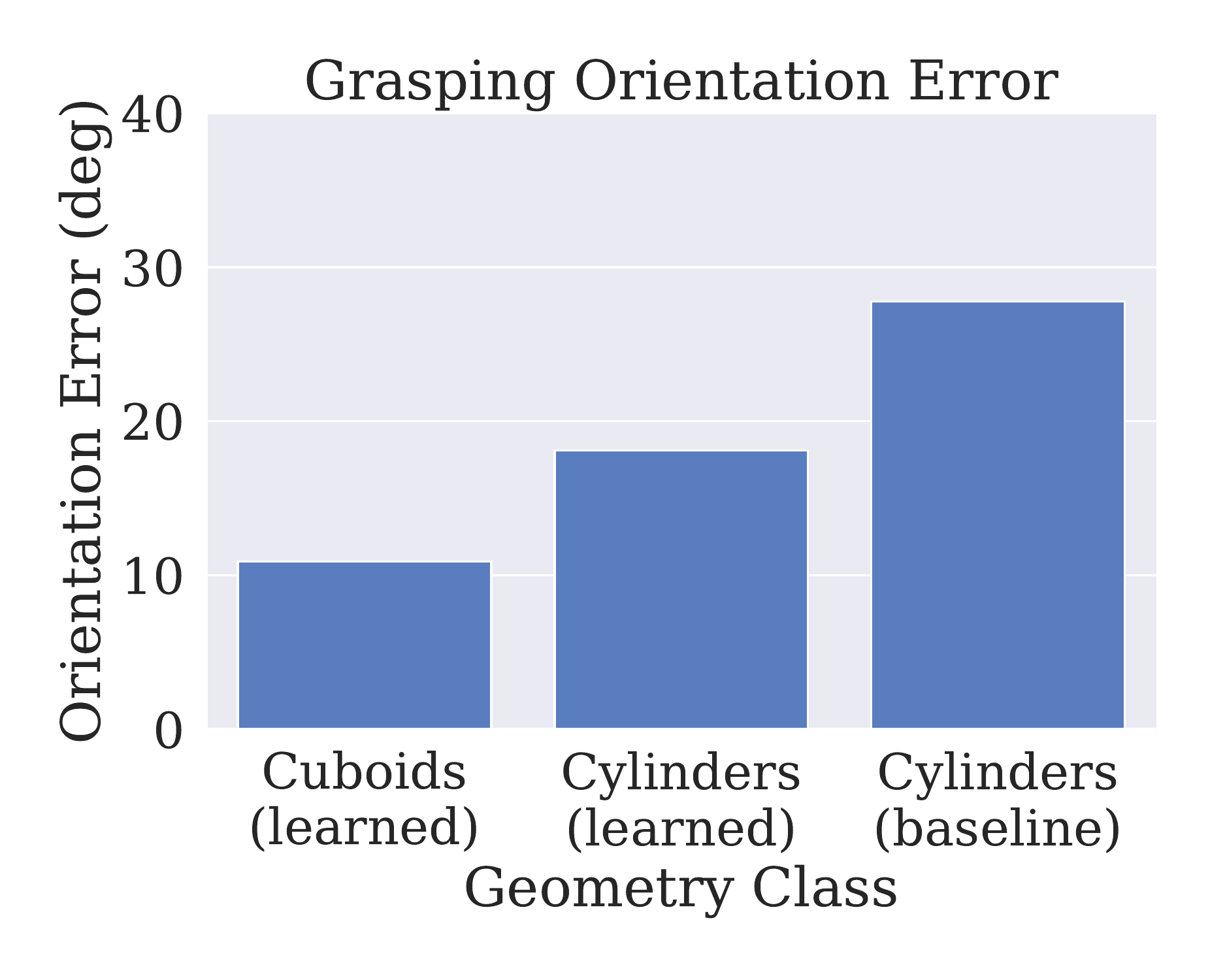}
		\caption{} 
		\label{fig:cuboid-cylinder-ori-error}
	\end{subfigure}\hfill
    \end{tabular}    
    \caption{Single-step \textit{grasp-reorient} results on novel cylindrical objects}
    \label{fig: cylinder-single-step-breakdown}
    \vspace{-10pt}
\end{figure}

%% file: figText/baseline_compare/failure_table.tex
\def\arraystretch{1.0}
\begin{table}[t]
\small
\centering
\caption{Comparing the fraction of failed planning attempts that were due to specific failure modes between learned and hand-designed samplers. 50 planning problems were setup with \texttt{P}$\shortrightarrow$\texttt{G}$\shortrightarrow$\texttt{P} plan skeleton, and different reasons for failure were tracked. Values denote the percent of overall trials where a particular failure mode occurred (note modes may overlap, so percentages don't add to 100).}
\begin{tabular}{c|c|c|c|c}
\hline \hline
Type & Colliding Start (\%) & Colliding Goal (\%) & Path Infeasible (\%) & Precondition (\%) \\ \hline 
Learned & 3.5 & 13.4 & 44.2 & 62.4 \\
Hand Designed & 7.3 & 38.9 & 70.5 & 43.4 \\
\hline \hline
\end{tabular}
\label{tab: baseline-compare-failure}
\end{table}

%% file: figText/baseline_compare/samples_table.tex
\def\arraystretch{1.0}
\begin{table}[t]
\small
\centering
\caption{Average number of sampling iterations performed during multi-step planning on a \texttt{P}$\shortrightarrow$\texttt{G}$\shortrightarrow$\texttt{P} skeleton over 50 trials, using the learned samplers and the hand-designed baseline. Total denotes the overall average over both successful and failed planning attempted, while Successes and Failures divides the average number of sampling iterations performed based on whether a feasible plan was found or not, respectively. Results highlight that both methods require similar computation (as indicated by the similar number of samples on failed attempts) and that the learned samplers are biased toward finding feasible parameters (as indicated by the lower number of samples required on successful attempts).} 
\begin{tabular}{c|c|c|c}
\hline \hline
Type & Total & Successes & Failures \\ \hline 
Learned & 46.5 & 37.6 & 137.7 \\
Hand Designed & 113.3 & 70.7 & 130.1 \\
\hline \hline
\end{tabular}
\label{tab: baseline-compare-samples}
\end{table}